\documentclass{article} 
\usepackage{iclr2026_conference,times}

\usepackage{amsmath,amsfonts,bm}

\def\eqref#1{equation~\ref{#1}}

\def\1{\bm{1}}

\DeclareMathAlphabet{\mathsfit}{\encodingdefault}{\sfdefault}{m}{sl}
\SetMathAlphabet{\mathsfit}{bold}{\encodingdefault}{\sfdefault}{bx}{n}

\newcommand{\R}{\mathbb{R}}

\newcommand{\softmax}{\mathrm{softmax}}

\DeclareMathOperator*{\argmax}{arg\,max}

\usepackage{url}
\newcommand{\ourmaintitle}{When Flatness Does (Not) Guarantee Adversarial Robustness}

\usepackage{comment}
\usepackage{latexsym}
\usepackage[ruled, vlined, nofillcomment, linesnumbered]{algorithm2e}
\usepackage[notheorems]{eda} 
\usepackage{bbm}
\usepackage{caption}
\usepackage{subcaption}
\usetikzlibrary{calc}
\usepackage{wrapfig}
\graphicspath{ {./figs/} }
\usepackage{prettyplots} 

\input{defines}
\newcommand{\pred}{ \hat{y} \xspace}
\newcommand{\partl}[1]{\frac{\partial}{\partial w_{#1}} \xspace}
\DeclareMathOperator{\pro}{\Pi}

\newcommand{\resnet}{\textsc{ResNet-18}\xspace}
\newcommand{\wrn}{\textsc{WideResNet-28-4}\xspace}
\newcommand{\dense}{\textsc{DenseNet121}\xspace}
\newcommand{\vgg}{\textsc{VGG11}\xspace}

\newcommand{\llama}{\textsc{LLama2}\xspace}
\newcommand{\vicuna}{\textsc{Vicuna}\xspace}
\newcommand{\guanaco}{\textsc{Guanaco}\xspace}

\usepackage{hyperref}
\usepackage{url}

\title{\ourmaintitle}

\author{%
  Nils Philipp Walter\\
  CISPA Helmholtz Center\\ for Information Security\\
  \texttt{nils.walter@cispa.de}\\
  \And
  Linara Adilova\\
  Ruhr University Bochum\\
  \texttt{linara.adilova@rub.de}\\
  \And
  Jilles Vreeken\\
  CISPA Helmholtz Center\\ for Information Security\\
  \texttt{vreeken@cispa.de}\\
  \And
  Michael Kamp\\
  Lamarr Institute, Technical University Dortmund \&\\
  Institute for AI in medicine, University Hospital Essen\\
  \texttt{michael.kamp@uk-essen.de} \\
}

\iclrfinalcopy 
\begin{document}

\maketitle

\begin{abstract}
   
Despite their empirical success, neural networks remain vulnerable to small, adversarial perturbations. A longstanding hypothesis suggests that flat minima, regions of low curvature in the loss landscape, offer increased robustness. While intuitive, this connection has remained largely informal and incomplete. By rigorously formalizing the relationship, we show this intuition is only partially correct: flatness implies \emph{local} but not \emph{global} adversarial robustness. To arrive at this result, we first derive a closed-form expression for relative flatness in the penultimate layer, and then show we can use this to constrain the variation of the loss in input space. This allows us to formally analyze the adversarial robustness of the entire network. We then show that to maintain robustness beyond a local neighborhood, the loss needs to curve \emph{sharply} away from the data manifold.
We validate our theoretical predictions empirically across architectures and datasets, uncovering the geometric structure that governs adversarial vulnerability, and linking flatness to model confidence: adversarial examples often lie in large, flat regions where the model is confidently wrong. Our results challenge simplified views of flatness and provide a nuanced understanding of its role in robustness.

\end{abstract}

\section{Introduction}
\label{sec:intro}

\begin{figure}[t]
\centering
\includegraphics[width=\textwidth]{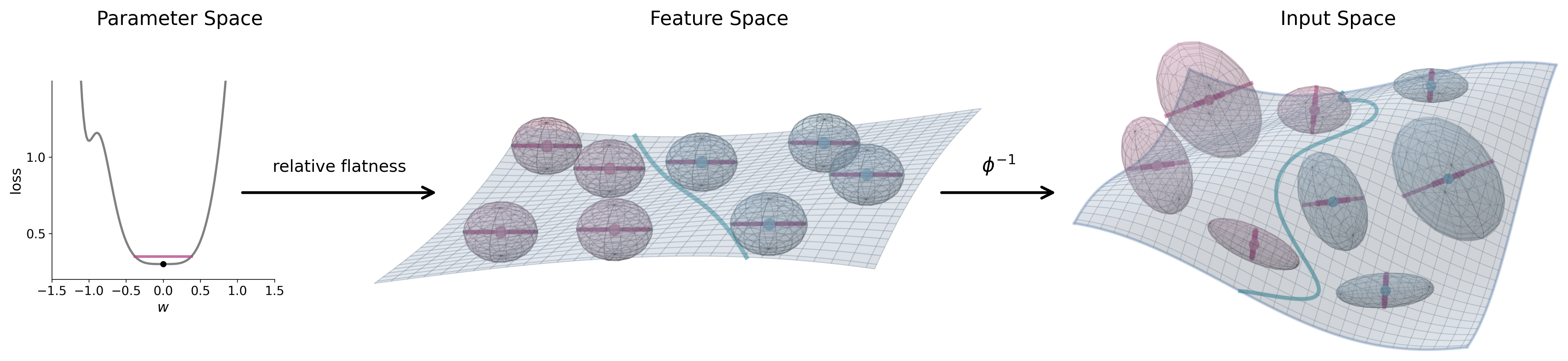}
\caption{
    \textbf{Flatness shapes robustness through the network.}
    Flatness in parameter space (left) implies a minimum robustness radius in feature space (center), which maps back to warped but bounded regions in input space (right). This illustrates how local flatness in the weight space of the penultimate layer translates into localized but non-uniform robustness in the input space.
    }
    \label{fig:flatness_input_feature} \vspace{-0.2cm}
\end{figure}

Despite their success across a wide range of tasks, neural networks remain notoriously brittle under adversarial perturbations. Small, often imperceptible changes to the input can dramatically alter a model's prediction. Understanding the structural properties that contribute to this vulnerability is central to building more robust systems.
One property that has long attracted attention is the flatness of the loss surface. Earlier work suggested that flatter minima correlate with better generalization~\citep{hochreiter1994simplifying, jiang2019fantastic}, however, the universality of this link remains an open question~\citep{andriushchenko2023modern}. Flatness also emerged as a potential indicator for adversarial robustness\citep{wu2020adversarial}: a model whose loss landscape is locally flat in parameter space might resist small perturbations in input space. At first glance, this appears to be disconnected, since adversarial examples concern the change of the loss with respect to the input, while flatness quantifies the change with respect to the weights. Still, in simple settings, one can observe a connection. For a linear model and a given training sample $x$, we have:
$$
\ell(f(\mathbf{w}(x + \delta x)), y) = \ell(f(\mathbf{w}x + \mathbf{w}\delta x), y) 
= \ell(f((\mathbf{w} + \mathbf{w}\delta)x), y) 
= \ell(f(\hat{\mathbf{w}}x), y) 
 \cong \ell(f(\mathbf{w}x), y)
$$
The last step follows from the loss surface being flat. That is, by definition the loss does not change under small perturbations. This informal derivation suggests that input perturbations can, in simple cases, be reinterpreted as weight perturbations. However, it oversimplifies the nonlinear structure of neural networks. In this work, we resolve this disconnect through a formal theoretical analysis. The main challenge lies in accounting for nonlinearity, which fundamentally alters how perturbations propagate through the network. To overcome this, we extend the theoretical results by~\citet{petzka2021relative}, who related a carefully defined notion of flatness in a single layer, based on the trace of the hessian $Tr(H)$, that binds the robustness to perturbations in feature space. 
Here, we show that their theoretical framework can be extended to relate the flatness of the loss surface to adversarial robustness. Intuitively, we show that under theory-friendly conditions, flatness in feature space constrains how much the loss can change in input space. This relationship is illustrated in Figure~\ref{fig:flatness_input_feature}: point-wise flatness, measured at individual inputs, corresponds to spherical regions in feature space where the loss changes slowly.
These map back to regions in input space---potentially warped or elliptical---where the loss also varies gradually, including under adversarial perturbations.
 
We then make this intuition precise. To this end, we derive a closed-form expression for relative flatness in the penultimate layer for the cross-entropy loss. This enables us to characterize how local curvature in parameter space influences the variation of the loss in feature space. By bounding the third derivative of the loss, we show that flatness at the penultimate layer induces a lower bound on the radius within which the loss remains approximately constant in feature space. Through the local Lipschitz continuity of the feature extractor, these spherical regions propagate back into the input space, where they become warped and anisotropic but retain their minimal safe diameter. Crucially, the size of these robust regions is dictated directly by the relative flatness at the penultimate layer.%

Our derivations and experiments confirm a long-suspected relationship between flatness and robustness; yet they expose its limitations. 
The theoretical picture is appealing but incomplete. While it is tempting to believe that a flatter model is automatically more robust, our results show that this can be misleading. Flatness ensures local robustness around individual points where it is measured, but it does not characterize the structure of the loss landscape globally. In particular, we observe that first-order adversarial attacks tend to move examples from initially sharp regions into much flatter ones. That is, adversarial examples often settle in flat basins of the loss, which we call the \emph{Uncanny Valley} (cf.~Fig.~\ref{fig:uncanny-exp}).
To understand this behavior, we revisit our closed-form expression for relative flatness in the penultimate layer. This expression reveals a tight connection between local geometry and model confidence: flatness tends to emerge in regions where the model is highly confident. In turn, this helps explain why adversarial examples can appear deceptively safe--despite being incorrect, they often reside in regions of low curvature and high confidence. This local flatness backpropagates through the network, influencing sensitivity in earlier layers. This theoretical lens enables us to rigorously invalidate overly simplified claims of the form: \emph{Flat minima result in adversarially more robust networks}.%

In summary, we make the following contributions:
\begin{enumerate}[label=(\roman*),itemsep=0pt, topsep=0pt, leftmargin=*]
\item We establish a theoretical connection between relative flatness and adversarial robustness and empirically validate these results.
\item We derive a closed-form expression for relative flatness in the penultimate layer, revealing its link to confidence and its influence on network sensitivity.
\item We introduce and analyze the \emph{Uncanny Valley}, showing that adversarial examples often lie in flat, vast, high-confidence regions despite being misclassified.
\end{enumerate}

These results provide a clearer understanding of how flatness and adversarial robustness interact. While flatness offers meaningful local guarantees, our theory reveals structural blind spots that still permit the existence of adversarial examples. This bridges parameter-space geometry and input-space robustness through both theoretical analysis and empirical evidence.

\section{Related Work}\label{sec:related}

\paragraph{Adversarial Examples}
\citet{szegedy2014intriguing} introduced the notion of adversarial samples. An adversarial sample is a subtle perturbation of a benign input that remains imperceptible to humans but causes a model to make incorrect predictions. 
Numerous methods for crafting adversarial examples have been developed. Early approaches include the Fast Gradient Sign Method (FGSM)~\citep{goodfellow2014explaining},
followed by more powerful and widely adopted techniques such as Projected Gradient Descent (PGD)~\citep{madry2017towards} and
the C\&W~\citep{carlini2017towards}. Further work expanded attack methods~\citep{papernot2016limitations, kurakin2017adversarial, narodytska_simple_2017, brown_adversarial_2018, alaifari2018adef, andriushchenko2020square, croce2020reliable, croce2021mind},
typically relying on gradient-based  optimization. In general, all deep learning based models, including large language models are vulnerable to input manipulations~\citep{li2020bert, zou2023universal, wei2024jailbroken}.
Because of many established attacks and smoothness of the input space we here focus on the image domain. Large language models are likewise vulnerable to input manipulations~\citep{li2020bert, zou2023universal, wei2024jailbroken}.
Here, we focus on adversarial in the image domain, due to the established attacks and smoothness of the input space.
In general, all deep learning based models, including large language models, are vulnerable to input manipulations~\citep{li2020bert, zou2023universal, wei2024jailbroken}.
Because of many established attacks and smoothness of the input space we here focus on the image domain. 

\paragraph{Flatness for Generalization \& Robustness }
The notion of flat minima was introduced by \citet{hochreiter1994simplifying}, who argued that such solutions correspond to simpler models with better generalization. This view gained empirical support from \citet{Keskar2017LargeBatch}, who showed that small-batch training tends to converge to flatter minima with lower test error, while large-batch training leads to sharper, less generalizable solutions. Subsequent studies linked generalization to sharpness metrics such as the Hessian trace and maximum eigenvalue~\citep{jiang2019fantastic, neyshabur2017exploring}.
These insights inspired a range of optimization techniques aimed at biasing training towards flatter regions of the loss surface. Methods like Entropy-SGD~\citep{chaudhari2019entropy}, Stochastic Weight Averaging (SWA)\citep{izmailov2018averaging}, SWAD \citep{cha2021swad}, and Sharpness-Aware Minimization (SAM)~\citep{foret2021sam} explicitly or implicitly minimize sharpness and have been shown to improve generalization across various architectures and datasets.

The relation between flatness and generalization remains topic of debate, however. \citet{dinh2017sharp} showed that sharpness measured via the loss Hessian is not invariant under reparameterization, undermining raw curvature-based metrics. \citet{kaurmaximum} and \citet{andriushchenko2023modern} found that sharpness often reflects training hyperparameters more than generalization quality, and may not reliably predict out-of-distribution performance. To address these issues, reparameterization-invariant definitions such as adaptive sharpness~\citep{kwon2021asam} and relative flatness~\citep{petzka2021relative} were proposed, with the latter also providing theoretical generalization guarantees. 
Flatness has also been studied in the context of adversarial robustness. \citet{stutz_relating_2021} analyzed curvature under $\ell_p$-bounded perturbations and observed that flatter input-space regions tend to correlate with robustness. Subsequent works showed that robust models often lie in flatter regions of the weight landscape~\citep{zhao2020bridging}, and that sharpness-aware training alone can enhance robustness even without adversarial examples~\citep{wu2020adversarial, wei2023sharpness}.

\section{Preliminaries}
\label{sec:prelim}

We assume a distribution $\D$ over an input space $\X$ and a label space $\Y$ with corresponding probability density function $P(X,Y) = P(Y\mid X)P(X)$. We consider models $f: \X \rightarrow \Y$ from a model class $\F$, and loss functions $l: \Y \times \Y \rightarrow \R_+$.

\subsection{Classic Definition of Adversarial Examples and Robustness}

An adversarial example is a minimal perturbation $r^*$ added to a sample $x$ such that $\xi = x + r^*$ is misclassified. 
Formally, it is defined as an optimization problem.
\begin{df}[Prediction-change adversarial example \citep{szegedy2014intriguing}]
Let $f: \R^m \rightarrow \{1,\dots,k\}$ be a classifier, $x \in [0,1]^m$, and $l \in [k]$ with $l = f(x)$ the predicted class. Then
\[
r^* = \arg\min_{r \in \R^m} \|r\|_2 \quad \text{s.t.} \quad f(x+r) \neq l \quad \text{and} \quad x+r \in [0,1]^m,
\]
and the perturbed sample $\xi = x + r^*$ is called an \textbf{adversarial example}.
\label{def:advEx-classic}
\end{df}

Conversely, a classifier is said to be adversarially robust at a sample $x$ if its predictions remain constant under small perturbations i.e. there exists no small perturbation such that the label flips.

\begin{df}[Prediction-change Pointwise Robustness]
A classifier $f$ is said to be \emph{$\delta$-robust} at $x \in \X$ if
$
\forall \xi \in B_\delta(x), \quad f(\xi) = f(x),
$
where $B_\delta(x) = \{ \xi \in \X \mid \|x - \xi\|_2 \leq \delta \}$ denotes the closed ball of radius $\delta$ around $x$.
\label{def:pointwiseRobust}
\end{df}
%
%
\begin{df}[Prediction-change Dataset Robustness]
Given a dataset $S \subseteq \X \times \Y$, a classifier $f$ is \emph{$\delta$-robust over $S$} if it is $\delta$-robust at every $(x,y) \in S$ according to Definition~\ref{def:pointwiseRobust}.
\label{def:datasetRobust}
\end{df}

\subsection{Loss-based Definitions of Adversarial Examples and Robustness}

In this work, we connect the geometry of the \emph{loss} surface with adversarial robustness. The classical definition of adversarial examples (Definition \ref{def:advEx-classic}) only incorporates the loss implicitly through the prediction flip, making it unsuitable for directly analyzing how robustness relates to the structure of the loss landscape. To address this, we adopt a straight-forward modification.

\begin{df}[Loss-change adversarial example ]
Let $f: \R^m \rightarrow \{1,\dots,k\}$ be a classifier, $x \in [0,1]^m$, and $l = f(x)$ the predicted class. Let $\ell: \{1,\dots,k\} \times \{1,\dots,k\} \to \R_+$ be a loss function, and let $\epsilon > 0$ be a threshold. Then
\[
r^* = \arg\min_{r \in \R^m} \|r\|_2 \quad \text{s.t.} \quad \ell(f(x+r), l) - \ell(f(x), l) > \epsilon \quad \text{and} \quad x+r \in [0,1]^m,
\]
and the perturbed sample $\xi = x + r^*$ is called a $(\delta, \epsilon)$-adversarial example, where $\delta = \|r\|_2$.
\label{def:advLossChange}
\end{df}
Instead of requiring a change in prediction, we require the loss to increase by more than a threshold $\epsilon$. This retains the core
idea of adversarial examples i.e. how small input perturbations have a malicious effect on the model, while capturing how perturbations in the input space translate to changes in the loss surface, rather than relying solely on discrete prediction flips.
This definition provides a generalization of the classic definition. Any adversarial example from Definition \ref{def:advEx-classic} satisfies Definition \ref{def:advLossChange} for some $\epsilon > 0$, and by using a conservative $\epsilon > \log(k)$ for cross-entropy loss, we can ensure a prediction-flip, which recovers the classical criterion. We make this precise in Lemma \ref{lem:adv-examples}. Analogous to Definition \ref{def:pointwiseRobust}, we define robustness against adversarial loss changes.

\begin{df}[Loss-change robustness]
Let $f: \X \to \Y$ be a classifier, $l: \Y \times \Y \to \R_+$ a loss function, and $(x,y) \in \X \times \Y$ a sample.  
Given thresholds $\delta > 0$ (perturbation radius) and $\epsilon > 0$ (loss tolerance),  
we say that $f$ is \emph{$(\delta, \epsilon)$-robust} at $(x,y)$ if
$
\forall \xi \in B_\delta(x), \quad l(f(\xi), y) - l(f(x), y) \leq \epsilon.
$
\label{def:lossRobustness}%
\end{df}\vspace{-0.35cm}%
In this setting, robustness requires that small perturbations do not substantially increase the loss. This makes robustness a continuous property, where the degree of robustness is controlled by the choice of $\epsilon$. A smaller $\epsilon$ enforces stricter stability, while a larger one allows more tolerance. For example, a perturbation may not flip the predicted label but can still reduce the model’s confidence, leading to a higher loss. The classical definition overlooks such cases, but the loss-based view captures such vulnerabilities naturally. We now extend the pointwise definition to the dataset level.

\begin{df}[Loss-change dataset robustness]
Given a dataset $S \subseteq \X \times \Y$, a classifier $f$ is \emph{$(\delta, \epsilon)$-robust over $S$} if it is $(\delta, \epsilon)$-robust at every $(x, y) \in S$ according to Definition~\ref{def:lossRobustness}.
\label{def:lossDatasetRobust}
\end{df}
It is easy to see that by choosing
$\epsilon_*(S; \delta) := \min_{(x,y) \in S} \inf_{\xi \in B_\delta(x): f(\xi) \neq y} \left[ l(f(\xi), y) - l(f(x), y) \right],
$
one recovers the prediction-flip criterion. This expression captures the smallest loss increase over all $\delta$-perturbations that
change the prediction. With this choice, no prediction flip will incur. This loss-based formulation is strictly stronger, though, as it requires small loss under perturbation even for high-confidence predictions---enforcing a stricter, uniform standard across the dataset.  
Alternatively, allowing $\epsilon$ to vary per sample recovers the original pointwise definition exactly and offers greater flexibility.
For ease of analysis, we focus on the stronger version with a fixed global $\epsilon$.

\section{What is Flatness Made Of?}
Flatness measures aim to quantify how sensitive a neural network's loss is to perturbations in parameters. Empirical approaches estimate flatness heuristically by applying random perturbations to model weights and observing the resulting loss variations~\citep{Keskar2017LargeBatch}. Analytical methods quantify curvature, using the trace or eigenvalue spectrum of the Hessian of the loss~\citep{hochreiter1994simplifying}. Both approaches, however, are expensive to compute in high-dimensional settings and offer limited support for analytical reasoning---especially eigenvalue-based measures, which are harder to manipulate symbolically and less amenable to theoretical analysis.

Among Hessian-based metrics, the trace is more practical and often used in theoretical analysis due to its simpler structure. However,
the raw trace is not invariant to reparameterization and thus does not reliably correlate with generalization~\citep{dinh2017sharp}.
To address these limitations, \citet{petzka2021relative} proposed a reparameterization-invariant flatness measure, \emph{relative 
flatness}, derived from a theoretical decomposition of the generalization gap. The measure combines the trace of the Hessian with the 
norm of the weights in a single layer. They show that, under mild technical assumptions, low relative flatness implies a small 
generalization gap. Importantly, they informally argue that it is sufficient to compute this quantity at the \emph{penultimate layer},
avoiding the need to analyze the full network. Below, we provide a formal justification for this claim.
Empirically, relative flatness also correlates strongly with generalization performance, making 
it both theoretically grounded and practically informative. 
Formally, given a network decomposed into a feature extractor $\phi: \mathcal{X} \rightarrow \mathbb{R}^d$ mapping to the penultimate layer $\mathbf{w} \in \mathbb{R}^{k \times d}$
and a classifier $g$ applying softmax, i.e., $f(x) = g(\mathbf{w} \phi(x))$ relative sharpness is defined as, 
\begin{equation}
    \kappa_{Tr}(\w) := \|w\|_2 Tr(H(\w,S))\enspace ,    
\end{equation}
where $Tr$ denotes the trace, and $H$ is shorthand for the Hessian of the loss computed on $S$ wrt $\w$:
\begin{equation}
H(\w,S)=\frac{1}{|S|}\sum_{(x,y)\in S}\left(\frac{\partial^2}{\partial w_i\partial w_j}\loss\left(g(\w\phi(x)),y\right)\right)_{i,j\in [km]}\enspace .
\end{equation}
This simplified measure~\citep{petzka2019reparameterization} upper bounds full relative flatness~\citep[Lm A.2][]{han2025flatness}.
Counter-intuitively, this measure is small when the loss surface is flat, since in that case the trace of the Hessian is small. Similarly, a large value of $\kappa_{Tr}(\w)$ indicates that the loss surface is sharp. To avoid confusion, we therefore call $\kappa_{Tr}(\w)$ \emph{relative sharpness}.

To understand the geometric structure of relative sharpness and enable the theoretical analysis in Section 4,
we need an explicit formula for the Hessian. We therefore derive a closed-form expression of the Hessian
for the cross-entropy loss, which reveals the underlying geometry and serves as the foundation for our theoretical
analysis. The full derivation is provided in Appendix~\ref{app:hessian}. For a single example $(x,y)$, defining the
softmax predictions as $\hat{y}=g(\w\phi(x))$, our derived expression is
\begin{equation}
H(\w,\{(x,y)\})  = (\text{diag} (\pred) - \pred \pred^T  ) \otimes \phi \phi^T \; 
\end{equation}
yielding a simple analytical formula,
\begin{equation}
    \kappa_{Tr}(\w) = \|w\|_2  tr(H(\w,\{(x,y)\})) = \|w\|_2 \sum_{j=1}^k \pred_j(1-\pred_j)\sum_{i=1}^d \phi_i^2 \; , \label{eq:simplifiedHessian}
\end{equation}
Thus, the geometry as measured by relative sharpness consists of three intuitive components:
the network's confidence (captured by $\hat{y}_j(1-\hat{y}_j)$), the scale of the feature representation
($\phi_i^2$), and the weight magnitude ($\|w\|_2$). Although this derivation explicitly applies to the penultimate
layer, we will demonstrate that this geometry propagates to earlier layers. This analytical form is key to
bounding adversarial susceptibility using relative sharpness in the following section.

\paragraph{Why the last layer is sufficient}
\label{sec:hesssuff}
For any recitified affine neural network composed of linear, convolutional and BatchNorm layers, curvature propagates cleanly
through the architecture. Let \(f_l(x) = \operatorname{ReLU}(W_l x)\) denote the \(l\)-th layer, with pre-activations
\(a^l = W_l x^{l-1}\). Define the gradient and Hessian of the loss \(\ell\) with respect to these pre-activations
as \(g^l = \partial \ell / \partial a^l\) and \(H^l = \partial^2 \ell / \partial (a^l)^2\). Since ReLU has zero second derivative almost everywhere, and each affine map introduces no additional curvature, the chain rule yields the recurrence  
\[
H^{l-1} = W_l^\top D^l H^l D^l W_l,
\]
where \(D^l = \mathrm{diag}(1_{a^l > 0})\) masks the active units. The Hessian block with respect to the weights of layer \(l\) takes the Kronecker-factored form  
\[
H_{W_l} = (x^{l-1} x^{l-1\top}) \otimes (D^l H^l D^l).
\]
If the final-layer Hessian
vanishes, so do all earlier activation- and weight-space blocks. Therefore, the curvature profile of the entire network is
fully determined by the last layer, and any sharpness measure based on the Hessian attains its maximum there. We formalize this
result in Appendix~\ref{app:hessian}.

This structure reveals that all curvature in hidden layers is inherited from the output layer. In particular,
if \(H^L = 0\), then by induction all downstream Hessians also vanish. Even when \(H^L \neq 0\), the conditioning of
every earlier block is shaped by it. As a result, any sharpness measure based on the Hessian is strongly influenced by
the final layer and its immediate input, justifying our focus on the penultimate layer in what follows. We formalize
this result in Appendix~\ref{app:hessianshape}.  \footnote{This does not hold for transformer since attention can introduce curvature.}

\section{Connecting Relative Sharpness to Adversarial Robustness}
\label{sec:theory}
In this section  we make explicit how relative sharpness influences adversarial robustness.  To achieve this,
we first connect input perturbations to changes in the output, then relate these changes in feature space to parameter perturbations, i.e., relative sharpness. Second, we use Taylor expansion around adversarial
examples to bound the loss increase under such perturbations. Finally, we derive the robustness bound that reveals
the role of relative sharpness. This precisely shows how relative sharpness influences the relationship between input perturbations and loss variations, though the resulting bound is not intended as practically tight guarantees.
As the derivations are technical, we present the key ideas and proof sketches here and
defer full details to the appendix.

\paragraph{Connecting input to feature space}

We first formalize how input-space perturbations propagate into feature-space representations. This is essential since neural
networks transform perturbations nonlinearly, complicating the input-to-parameter relationship. Formally, consider a model
\( f(x) = g(\mathbf{w}\phi(x)) \) with feature extractor \(\phi\), classifier \(g\), and weight matrix \(\mathbf{w}\).
Such a model nonlinearly relates input perturbations to those in feature space~\citep{petzka2021relative}. Specifically,
we represent the adversarial representation \(\phi(\xi)\) as a perturbation of the clean representation \(\phi(x)\).
\begin{restatable}{lm}{relationFeatureInput}
    Let $f=g(\w\phi(x))$ be a model with $\phi$ $L$-Lipschitz and $\|\phi(x)\|\geq r$, and $\xi,x\in\X$ with $\|\xi - x\|\leq \delta$, then there exists a $\Delta>0$ with
    $\Delta \leq L\delta r^{-1}$, such that $\phi(\xi)=\phi(x)+\Delta A\phi(x)$, where $A$ is an orthogonal matrix.
\label{lm:relationFeatureInput}
\end{restatable}
The proof is provided in Appx.~\ref{appdx:proofLmRelationFeatureInput}. 
Thus we can relate the adversarial example $\xi$ to perturbations in the weights $\w$ of the representation layer, for which we use the linearity argument from~\citet{petzka2021relative}. Since we can express $\phi(\xi)$ as $\phi(x)+\Delta A\phi(x)$, we can then use the linearity of the representation layer to relate the perturbation of input to a perturbation in weights. That is,
\[
\loss(f(\xi),y)=\loss(g(\w\phi(\xi)),y)=\loss(g(\w(\phi(x)+\Delta A\phi(x))),y)=\loss(g((\w+\Delta \w A)\phi(x),y)\enspace .
\]
This means that we can express an adversarial example as a suitable perturbation of the weights $\w$ of the representation layer, where the magnitude is bounded by $\Delta\leq L\delta r^{-1}$.

\paragraph{Bounding loss increase}
Next, we  bound the loss difference between $\xi$ and $x$. For ease of notation, we define $\loss(\w+\Delta \w A):=\loss(g((\w+\Delta \w A)\phi(x),y)$. The Taylor expansion of $\loss(\w+\Delta \w A)$ at $\w$ gives the following
\[
\loss(\w+\Delta \w)=\loss(\w)+\innerprod{\Delta \w A}{\nabla_\w\loss(\w)}+\frac{\Delta^2}{2}\innerprod{\w A}{H\loss(\w)(\w A)}+R_2(\w,\Delta)\enspace ,
\]
where $H\loss(\w)$ is the Hessian of $\loss(\w)$. If we now maximize over all $A$ with $\|A\|\leq 1$, it follows that  
$\innerprod{\w A}{H\loss(\w)(\w A)}\leq \|\w\|_F^2 Tr(H\loss(\w)) = \kappa^\phi_{Tr}(\w)$.
Therefore, we have
\begin{equation}
\left|\loss(f(\xi),y) - \loss(f(x),y)\right|\leq \Delta\|\w\|_F\|\nabla_\w\loss(\w)\|_F + \frac{\Delta^2}{2}\kappa^\phi_{Tr}(\w) + R_2(\w,\Delta)\enspace .
\label{eq:lossBoundTaylor}
\end{equation}
The remainder depends on the partial third derivatives of the loss. We show in Appdx.~\ref{app:lossBoundAdversarial} that for feature extractor $\phi$ that is $L$-Lipschitz, it can be bound by $4^{-1}kmL^3$.
With this we can bound the difference between the loss suffered on an adversarial example $\xi$ and the loss on a clean example $x$ for a converged model as follows.
\begin{restatable}{pr}{lossBoundAdversarial}
For $(x,y)\in\X\times\Y$ with $\|x\|\leq 1$ for all $x\in\X$, a model $f(x)=g(\w\phi(x))$ at a minimum $\w\in\R^{m\times k}$ with $\phi$ $L$-Lipschitz and $\|\phi(x)\|\geq r$, and the cross-entropy loss $\loss(\w)=\loss(g(\w \phi(x)),y)$ of $f$ on $(x,y)$, it holds for all $\xi\in\X$ with $\|x-\xi\|_2\leq\delta$ that
\begin{equation*}
    \begin{split}
        \loss(f(\xi),y) - \loss(f(x),y)\leq& \frac{\delta^2}{2r^2}L^2\kappa^\phi_{Tr}(\w) + \frac{\delta^3}{24r^3}kmL^6\enspace .
    \end{split}%
\end{equation*}%
\label{prop:lossBoundAdversarial}%
\end{restatable}\vspace{-0.5cm}%
We defer the proof to Appx.~\ref{app:lossBoundAdversarial}.
This analytical result clarifies explicitly how relative sharpness controls loss sensitivity to perturbations. 

\paragraph{Bounding adversarial robustness}
We now derive an explicit expression for the maximum perturbation radius $\delta$ within which the loss increase remains bounded by a given threshold $\epsilon$ . Solving the Taylor-based bound for $\delta$ yields a cubic equation with both linear and quadratic terms. While this equation does not admit a clean closed-form solution in general, we isolate and present the dominant terms to illustrate the key dependencies:
\begin{restatable}{pr}{adversarialRobustness}[Informal] \label{prop:bound}
For a dataset $S\subset\X\times\Y$ with $\|x\|\leq 1$ for all $x\in\X$, a model $f(x)=g(\w\phi(x))$ at a minimum $\w\in\R^{m\times k}$ wrt. $S$, with $\phi$ $L$-Lipschitz and $\|\phi(x)\|\geq r$, and the cross-entropy loss $\loss(\w)=\loss(g(\w \phi(x)),y)$ of $f$ on $(x,y)$, $d$ being the $L2$-distance, and $\epsilon>0$, $f$ is $(\epsilon,\delta,S)$-robust against adversarial examples with%
\[
\begin{aligned}
   \delta &\;\propto\;
   \frac{\epsilon^{\frac13}}{L\,\kappa_{Tr}(\w)^{\frac13}}
   \;+\;
   \frac{r\,k\,m\,L^{2}}{\kappa_{Tr}(\w)}
   \quad\quad
   &
   \lim_{\kappa_{Tr}(\w)\to 0}\delta &=
   \frac{r}{L}\left(\frac{24\,\epsilon}{kmL^{3}}\right)^{\tfrac{1}{3}}
\end{aligned}
\]
Hence decreasing sharpness  enlarges the certified radius,
yet the bound \emph{does not blow up}; it saturates at the value above. \label{prop:adversarialRobustness}
\end{restatable}


The formal version and full proof are provided in Appendix~\ref{appdx:proofPropAdvRobustness}. Proposition~\ref{prop:bound}  precisely reveals how relative sharpness constrains the extent to which the loss can change under input perturbations. Beyond its analytical role, it aligns with geometric intuition: flatter models in the sense of lower $\kappa_{Tr}(\w)$, together with smooth feature maps, exhibit locally stable behavior. 
Interestingly, the explicit formula for relative sharpness in Equation~\ref{eq:simplifiedHessian} reveals that sharpness \emph{decreases} as model confidence \emph{increases}. When the predicted class probability approaches one, the corresponding term $\pred_j(1-\pred_j)$ vanishes, reducing the overall sharpness. This implies that highly confident predictions are, at least \emph{locally}, more robust to input perturbations than uncertain ones. This stands in contrast to the common claim that high-confidence predictions are cause for adversarial susceptibility. Instead, our analysis suggests that high-confidence regions may actually promote local stability by flattening the loss surface. We revisit and empirically validate this observation in Section~\ref{sec:exps}.

\section{From Theory to Practice}
\label{sec:exps}
In this section, we empirically investigate how relative sharpness influences adversarial robustness. To test the theoretical predictions
from Section 4, we exploit a key insight: by scaling the penultimate layer weights $\mathbf{w}$ by a factor $s$ i.e. $\mathbf{w}_s = s\mathbf{w}$,
we can directly control the loss surface curvature without retraining. Smaller scaling factors yield sharper networks, while larger values
produce flatter ones. This method allows us to precisely control sharpness in a systematic
and reproducible manner, in contrast to SAM-based training approaches.

We design three experiments to examine different aspects of the sharpness-robustness connection. First, we examine whether the
curvature properties at the penultimate layer propagate throughout the network, validating our derivations in Section \ref{sec:hesssuff} and our focus on
this layer as representative of the global loss landscape. Second, we test our theoretical insight from Proposition 9
by examining whether networks with reduced sharpness demonstrate smaller loss changes when subjected to adversarial
perturbations, directly validating whether flatter networks exhibit improved stability under input perturbations.
Third, we investigate potential limitations of using relative sharpness as a global robustness and correctness indicator given its close connection to model confidence.

\paragraph{Setup}
We evaluate our theoretical predictions on standard architectures and datasets. We train ResNet-18 \cite{he2016deep}, WideResNet-28-4 \cite{Zagoruyko2016WideRN}, DenseNet-121 \cite{huang2017densely}, and VGG-11 with BatchNorm \cite{Simonyan2014VeryDC} on both CIFAR-10 and CIFAR-100. All models are trained using stochastic gradient descent for 100 epochs with an initial learning rate of 0.1, cosine learning rate scheduling, and weight decay of $10^{-4}$.
For clarity of presentation, we focus our main analysis on ResNet-18 trained on CIFAR-10. Our results are consistent across all architectures and datasets; hence we report on this standard combination. Complete results for other architectures and CIFAR-100 are provided in Appendix \ref{app:add-results}.

\subsection{Sharpness Predicts Loss Sensitivity Under Adversarial Perturbations}
\label{sec:eval}
We now empirically test our core claim that relative sharpness governs local loss sensitivity and shapes adversarial robustness. By scaling the network post-hoc,
we isolate the effect of sharpness while keeping all model parameters fixed, enabling controlled investigation of its role.
While flattening the network does not eliminate adversarial examples entirely, it significantly affects how rapidly the loss increases in their vicinity.
Specifically, we examine how local sharpness influences the magnitude of loss increase as defined in Definition~\ref{def:advLossChange}.

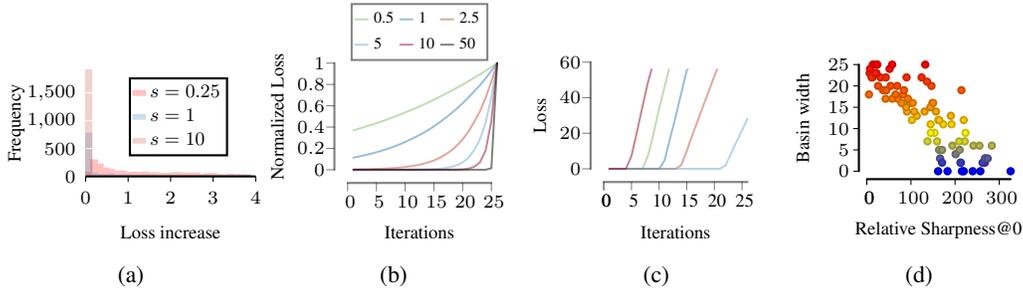
\begin{figure}[h!]
    \begin{subfigure}[b]{0.25\linewidth}
        \begin{tikzpicture}
            \begin{axis}[
                xlabel={Loss increase},
                ylabel={Frequency},
                pretty ybar,
                cycle list name = prcl-ybar,
                height=3cm,
                width=1.1\linewidth,
                xmin=0, xmax=4,
                xtick={0,1,2,3,4},
                legend style={yshift=-0.1cm,,xshift=-0.8cm,font=\scriptsize},
                legend style={font=\tiny},
                legend entries = {$s=0.25$,$s=1$,$s=10$},
                legend columns = 1,
                ylabel style={at={(axis description cs:-0.3,.5)},anchor=south},
                yticklabel style={xshift=2pt},
            ]

            \addplot +[
                hist={
                    bins=30,
                    },opacity = 0.25, red,
                    x filter/.code={
                    \pgfmathparse{(\pgfmathresult < 4) ? \pgfmathresult : nan}%
                    \let\pgfmathresult=\pgfmathresult},
            ] table [y index=1,col sep=comma] {new_results/losses/scale_0.25_resnet-eps-0_analysis_results.csv};

            \addplot +[
            hist={bins=30},
            opacity=0.25,
            %
            x filter/.code={
                \pgfmathparse{(\pgfmathresult < 4) ? \pgfmathresult : nan}%
                \let\pgfmathresult=\pgfmathresult
            },
            ]%
            table[y index=1, col sep=comma] {new_results/losses/scale_1_resnet-eps-0_analysis_results.csv};

            \addplot +[
            hist={bins=30},
            opacity=0.25,
            %
            x filter/.code={
                \pgfmathparse{(\pgfmathresult < 4) ? \pgfmathresult : nan}%
                \let\pgfmathresult=\pgfmathresult
            },
            ]%
            table[y index=1, col sep=comma] {new_results/losses/scale_10_resnet-eps-0_analysis_results.csv};

        \end{axis}
        \end{tikzpicture}
        \subcaption{}
        \label{sub:lossincrease-main}
    \end{subfigure}%
    \begin{subfigure}[b]{0.25\linewidth}
            \begin{tikzpicture}[baseline=(current axis.outer south)]
                    \usetikzlibrary{calc}
                    \begin{axis}[ 
                        smalljonas line,
                        cycle list name = prcl-line,
                        clip=false, 
                        legend style={
                            at={(0.5,1.02)}, anchor=south,    
                            font=\scriptsize
                          },
                        legend image post style={xscale=0.55},
                        width=\linewidth,
                        height=3cm,
                        ylabel={Normalized Loss},
                        xlabel={Iterations},
                        pretty labelshift,
                        line width=0.75,
                        xtick={0,5,10,15,20,25},
                        ymax=1,
                        ymin=0.0,xmin=0.0,xmax=25,
                        legend columns = 3,
                        tick label style={/pgf/number format/fixed},
                        legend entries = {Adv, Clean}
                    ]
    
                    \pgfplotsinvokeforeach{0.5,1,2.5,5,10,50}{%
                            \addplot+[
                                mark=*,
                            ] table[
                                y = normalized_losses,   
                                col sep = comma,
                                x expr = \coordindex + 1 
                            ] {new_results/losses/basin/#1_values.csv};
                            \addlegendentry{#1}          
                        }%

                    \end{axis}
                \end{tikzpicture}
                \subcaption{}
                \label{sub:basin}
    \end{subfigure}%
    \begin{subfigure}[b]{0.25\linewidth}
        \begin{tikzpicture}
                \usetikzlibrary{calc}
                \begin{axis}[ 
                    smalljonas line,
                    cycle list name = prcl-line,
                    legend style={yshift=-0.4cm,,xshift=-0.8cm,font=\scriptsize},
                    width=\linewidth,
                    height=3cm,
                    ylabel={Loss},
                    xlabel={Iterations},
                    pretty labelshift,
                    line width=0.75,
                    restrict y to domain*=0:60,
                    ymax=60,
                    ymin=0.0,
                    legend columns = 1,
                    xtick={0,0,5,10,15,20,25},
                    xticklabels =  {0,0,5,10,15,20,25},xmin=0,
                    tick label style={/pgf/number format/fixed},
                ]
                \pgfplotsinvokeforeach{loss_340,loss_541,loss_495,loss_581,loss_896}{
                    \addplot+[
                        mark=*,
                    ] table[
                        y = #1,   
                        col sep = comma,
                        x expr = \coordindex + 1 
                    ] {new_results/losses/takeoff_samples.csv};         
                }
                \addplot+[
    domain=1:26,  
    samples=2,     
    color=white,
    line width=5pt, opacity=1 
] {101};
                \end{axis}
            \end{tikzpicture}
                    \subcaption{}
                    \label{fig:takeoff}
    \end{subfigure}%
    \begin{subfigure}[b]{0.25\linewidth}
        \begin{tikzpicture}
                \usetikzlibrary{calc}
                \begin{axis}[ 
                    pretty scatter,
                    legend style={yshift=-0.4cm,,xshift=-0.8cm,font=\scriptsize},
                    width=\linewidth,
                    height=3cm,
                    ylabel={Basin width},
                    xlabel={Relative Sharpness@0},
                    pretty labelshift,
                    line width=0.75,
                    ymin=0, ymax=25,
                    legend columns = 1,
                    ytick={0,5,10,15,20,25},
                    xtick={0,0,100,200,300},
                    xmin=0,
                    tick label style={/pgf/number format/fixed},
                ]
                    \addplot[only marks, mark size=1.2pt]
                    table[x={sharpness_at_zeros}, y={take_off}, col sep=comma]{new_results/losses/resnet-eps-0/take_off_resnet-eps-0_analysis_results.csv};
                \end{axis}
            \end{tikzpicture}
                    \subcaption{}
                    \label{fig:takeoff-correlation}
\end{subfigure}%
    
    \caption{\textit{Loss geometry.} \textbf{(a)} We report the distribution of the loss increase for varying scaling values.
    \textbf{(b)} We show for one example how the basin formes as we increase the scaling. We use the normalized loss to plot all in one axis.
    \textbf{(c)} We show examples of samples exhibiting different basin widths.
    \textbf{(d)} We plot the sharpness measured at the test data ($s=1$) and the basin width. }
    \label{fig:losses}
    \end{figure}

\textbf{Setup.} To this end, we generate adversarial examples using a weaker attack PGD-$\ell_2$ (25 steps, $\epsilon = 0.025$, step size $\alpha = 0.001$).
This attack incrementally increases perturbations by 0.001$\ell_2$ per iteration, achieving a robust test accuracy of 90.33\% on the original model.
We evaluate scaled networks with scaling factors  $s \in \{0.25,0.5,1,2.5,5,10,50\}$ along the trajectories of these attacks,
measuring how the loss evolves. Results are summarized in Figure~\ref{fig:losses}, with a subset of scaling factors presented for clarity.
Comprehensive results for all scaling values are provided in the appendix.

\textbf{Loss Increase Analysis.} Figure~\ref{sub:lossincrease-main} shows the histogram of loss increases between clean examples $x_0$ and final adversarial iterates
$x_{25}$. As predicted by Equation~\ref{eq:lossBoundTaylor}, increasing the scale factor (flattening the network) substantially reduces observed loss increases,
approaching zero for large scaling values. This occurs because flattening creates locally stable regions--or basins--around inputs.

\textbf{Sharpness Creates Basins.} Figure~\ref{sub:basin} illustrates this phenomenon for a representative example. As we increase the sclaing $s$, the loss remains nearly
flat over progressively larger distances along the attack trajectory. For small scaling factors (e.g., $s = 0.5$), the loss increases steadily from the first iteration,
while for large factors (e.g., $s = 50$), the loss stays almost constant throughout the entire trajectory until a sharp \emph{take-off point} where it suddenly strongly increases,
which marks the boundary of the basin. Specifically, flatter loss surfaces, as derived in Proposition \ref{prop:adversarialRobustness}, 
can progressively guarantee larger adversarial robustness radii $\delta$ for given $\epsilon$, but only up to the take-off point, which represents the effective limit of sharpness when exclusively enforced by reducing $tr(H)$.

\textbf{Per Sample Basin.} The location of the take-off point varies across samples, as shown in Figure~\ref{fig:takeoff}.
In other words, the width of the basin is sample-dependent, and consequently, so is the guaranteeable radius. Following
conventional wisdom in the field, we observe that basin width strongly correlates with the sharpness measured at the clean
input--specifically, flatter networks (lower sharpness) yield broader basins. This relationship is empirically
confirmed in Figure~\ref{fig:takeoff-correlation}. This observation aligns with the established intuition that reduced
sharpness increases robustness, though within a radius that remains insufficient to prevent label flips.

\subsection{Sharpness Can Be Deceiving}
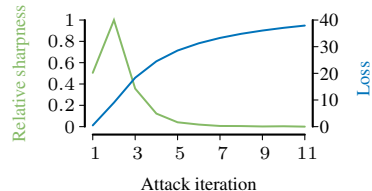
\begin{wrapfigure}[10]{r}{0.35\textwidth}
\vspace{-1.21cm}
    \begin{tikzpicture}
        \usetikzlibrary{calc}
        \begin{axis}[ 
            pretty line,
            legend style={yshift=1.0cm,font=\scriptsize,xshift=1cm},
            width=0.9\linewidth,
            height=3cm, 
            ylabel={Relative sharpness},
            xlabel={Attack iteration},
            pretty labelshift,
            line width=0.75,
            xtick = {1, 3, 5, 7, 9, 11},
            tick label style={/pgf/number format/fixed},
            axis y line* = left,
            axis x line* = bottom,
            ymin=0,
            ylabel style={text=dollarbill}
        ]
        \foreach \x in {wrn}{%
            \addplot[draw=dollarbill] table[ x={0}, y={trace-\x-cifar10}, col sep=comma]{expres/normalized_rfm-for-clean.csv};
        }    
        \end{axis}

        \begin{axis}[ 
            pretty line,
            width=0.9\linewidth,
            height=3cm,
            ylabel={Loss},
            pretty labelshift,
            line width=0.75,
            xtick = {1, 3, 5, 7, 9, 11},
            ytick = {0, 10, 20, 30, 40},
            yticklabels = {0,10,20,30,40}, 
            ymin=0,
            ymax=40,
            axis y line* = right,
            axis x line=none, 
            tick label style={/pgf/number format/fixed},
            ylabel style={text=frenchblue, xshift=-2pt},
            yticklabel style={xshift=-5pt}
        ]
        \foreach \x in {wrn}{%
            \addplot+[draw=frenchblue] table[ x={0}, y={loss-\x-cifar10}, col sep=comma]{expres/normalized_rfm-for-clean.csv};%
        }    
        \end{axis}
    \end{tikzpicture}
    \caption{\textit{Uncanny Valley}. On the PGD-trajectory, sharpness quickly peaks and then decreases to zero, while loss keeps increasing.}
    \label{fig:uncanny-exp}
\end{wrapfigure}%
In the previous experiments, we validated our theoretical derivations showing that relative sharpness governs local loss sensitivity, with flatter networks exhibiting limited loss variation under perturbations. We now turn to a more nuanced question. The flatness literature carries an implicit hope that geometric properties can be linked to correctness--that flatter minima naturally correspond to regions where the model generalizes better and makes accurate predictions. This hope is the implicit motivation of seeking flatter minima. To investigate this, we measure relative sharpness (the same holds for common classification losses, see Lemma \ref{lem:poly_growth_vanishes} \& Lemma \ref{lem:quad_pole}) along PGD-$\ell_\infty$ attack trajectories (10 steps, $\epsilon = 8/255$, $\alpha = 2/255$) to examine the evolving geometry.

We plot the results in Fig~\ref{fig:uncanny-exp}, which reveals a striking phenomenon: as the adversarial attack progresses, sharpness initially increases as expected. It then, however, dramatically decreases to near-zero values while the loss plateaus. This demonstrates that adversarial examples settle in exceptionally flat regions.
Our closed-form expression 
for relative sharpness explains this \emph{uncanny valley}; in high-confidence regions, confidence dominates 
the geometry, creating flat valleys where gradients nearly vanish. This challenges the hope that flatness indicates correctness--these regions are geometrically stable yet correspond to confident misclassifications.
The sharp peak in relative sharpness marks the decision boundary, where model uncertainty creates maximum curvature. The adversarial attack traverses this ridge before descending into flat, high-confidence regions on the wrong side. This reveals how confidence shapes the loss surface geometry---sharp at decision boundaries, flat in confident regions---\emph{regardless} of correctness.

\section{Discussion \& Conclusion}
\label{sec:discussion}

In this work, we formalized the long-suspected connection between flatness and adversarial robustness,
revealing both the promise and fundamental limitations of sharpness as a mediator for robustness. Our
analysis bridges parameter-space and input-space perspectives through rigorous theoretical derivations 
and extensive empirical validation. Traditional prediction-flip definitions of adversarial examples proved
insufficient for this analysis, so we introduced a loss-based formulation (Definition 4) that enables
direct analysis of how perturbations propagate through the loss landscape. This approach connects the
geometric properties at the penultimate layer to adversarial vulnerability in input space. We extended
\citep{petzka2021relative} framework to the adversarial setting, deriving explicit bounds (Proposition 9)
that show how relative flatness constrains the adversarial perturbation radius on the points it is measured on. The robustness guarantee
scales as $\epsilon^{1/3}/\kappa_{Tr}(w)^{1/3}$, providing a rigorous characterization of parameter-space geometry
influences adversarial robustness.

Our empirical investigations confirmed these theoretical predictions while exposing the myopic nature of flatness-based
robustness. Through controlled  experiments, we showed that while flatter networks exhibit larger stability basins 
around individual inputs, these basins remain insufficient for practical adversarial robustness. The basins have finite
width with sharp boundaries that act as "slingshots," rapidly increasing the loss once crossed.
We observe a string phenomenon; starting from relatively flat regions, attacks traverse increasingly sharp areas
before descending into exceptionally flat valleys. The peak corresponds to the decision boundary where predictions flip i.e 
maximal uncertainty. Beyond lie broad broad flat regions, termed the Uncanny Valley, where models are maximally confident yet wrong. This shows that confidence manifests in the 
geometry of the loss surface and that sharpness can be deceiving.

Our decision to analyze the network in two parts---feature extractor and classifier---proved crucial for understanding the
true nature of flatness in neural networks. By focusing our theoretical analysis on the penultimate layer, we could isolate
how classifier confidence dominates standard flatness measures. This decomposition revealed that $\text{Tr}(H)$-based metrics
conflate two different properties: the geometric stability of learned representations and the confidence of the
final classifier. When measuring flatness over the entire network, high confidence in the classifier
can mask the true sensitivity of the feature extractor. Our approach demonstrates that
meaningful robustness evaluation requires separately assessing these components. While the penultimate layer analysis might
initially seem like a limitation, it actually enabled us to uncover how confidence effects propagate through the network and
shape the global loss landscape.

In conclusion, by extending relative flatness to the adversarial setting, we have derived explicit
bounds that confirm flatness does influence robustness—but only through local mechanisms that cannot extend globally.
Our work demonstrates that flatness provides robustness within local basins whose size is fundamentally constrained by
the confidence-flatness coupling. As long as flatness measures at the penultimate layer are dominated by the confidence
term $\hat{y}_j(1-\hat{y}_j)$, they cannot distinguish between true geometric robustness and mere confident predictions.
This helps explain why optimizing for flatness, while beneficial, has proven insufficient for achieving strong adversarial
robustness. While $\text{Tr}(H)$-based flatness works as a local robustness indicator, achieving global robustness requires
developing new notions of flatness that meaningfully separate the contributions of the feature extractor from classifier confidence.

\bibliography{paper.bib}

\providecommand{\noopsort}[1]{}
\begin{thebibliography}{81}
\providecommand{\natexlab}[1]{#1}
\providecommand{\url}[1]{\texttt{#1}}
\expandafter\ifx\csname urlstyle\endcsname\relax
  \providecommand{\doi}[1]{doi: #1}\else
  \providecommand{\doi}{doi: \begingroup \urlstyle{rm}\Url}\fi

\bibitem[Alaifari et~al.(2018)Alaifari, Alberti, and
  Gauksson]{alaifari2018adef}
Rima Alaifari, Giovanni~S Alberti, and Tandri Gauksson.
\newblock Adef: an iterative algorithm to construct adversarial deformations.
\newblock In \emph{International Conference on Learning Representations}, 2018.

\bibitem[Alayrac et~al.(2019)Alayrac, Uesato, Huang, Fawzi, Stanforth, and
  Kohli]{alayrac2019labels}
Jean-Baptiste Alayrac, Jonathan Uesato, Po-Sen Huang, Alhussein Fawzi, Robert
  Stanforth, and Pushmeet Kohli.
\newblock Are labels required for improving adversarial robustness?
\newblock \emph{Advances in Neural Information Processing Systems}, 32, 2019.

\bibitem[Andriushchenko et~al.(2020)Andriushchenko, Croce, Flammarion, and
  Hein]{andriushchenko2020square}
Maksym Andriushchenko, Francesco Croce, Nicolas Flammarion, and Matthias Hein.
\newblock Square attack: A query-efficient black-box adversarial attack via
  random search.
\newblock In \emph{European Conference on Computer Vision}, pp.\  484--501,
  2020.

\bibitem[Andriushchenko et~al.(2023)Andriushchenko, Croce, M{\"u}ller, Hein,
  and Flammarion]{andriushchenko2023modern}
Maksym Andriushchenko, Francesco Croce, Maximilian M{\"u}ller, Matthias Hein,
  and Nicolas Flammarion.
\newblock A modern look at the relationship between sharpness and
  generalization.
\newblock In \emph{International Conference on Machine Learning}, pp.\
  840--902. PMLR, 2023.

\bibitem[Becker \& LeCun(1989)Becker and LeCun]{becker1989improving}
Sue Becker and Yann LeCun.
\newblock Improving the convergence of back–propagation learning with
  second–order methods.
\newblock In David~S. Touretzky, Geoffrey~E. Hinton, and Terrence~J. Sejnowski
  (eds.), \emph{Proceedings of the 1988 Connectionist Models Summer School},
  pp.\  29--37, San Francisco, CA, 1989. Morgan Kaufmann.

\bibitem[Bishop(1992)]{bishop1992exact}
Christopher~M. Bishop.
\newblock Exact calculation of the hessian matrix for the multilayer
  perceptron.
\newblock \emph{Neural Computation}, 4\penalty0 (4):\penalty0 494--501, 1992.
\newblock \doi{10.1162/neco.1992.4.4.494}.

\bibitem[Brown et~al.(2018)Brown, Mané, Roy, Abadi, and
  Gilmer]{brown_adversarial_2018}
Tom~B. Brown, Dandelion Mané, Aurko Roy, Martín Abadi, and Justin Gilmer.
\newblock Adversarial {Patch}, May 2018.
\newblock NIPS 2017, Long Beach, CA, USA.

\bibitem[Cai et~al.(2018)Cai, Liu, and Song]{cai2018curriculum}
Qi-Zhi Cai, Chang Liu, and Dawn Song.
\newblock Curriculum adversarial training.
\newblock In \emph{Proceedings of the 27th International Joint Conference on
  Artificial Intelligence}, pp.\  3740--3747, 2018.

\bibitem[Carlini \& Wagner(2017)Carlini and Wagner]{carlini2017towards}
Nicholas Carlini and David Wagner.
\newblock Towards evaluating the robustness of neural networks.
\newblock In \emph{2017 ieee symposium on security and privacy (sp)}, pp.\
  39--57. IEEE, 2017.

\bibitem[Carmon et~al.(2019)Carmon, Raghunathan, Schmidt, Duchi, and
  Liang]{carmon2019unlabeled}
Yair Carmon, Aditi Raghunathan, Ludwig Schmidt, John~C Duchi, and Percy~S
  Liang.
\newblock Unlabeled data improves adversarial robustness.
\newblock \emph{Advances in neural information processing systems}, 32, 2019.

\bibitem[Cha et~al.(2021)Cha, Chun, Lee, Cho, Park, Lee, and Park]{cha2021swad}
Junbum Cha, Sanghyuk Chun, Kyungjae Lee, Han-Cheol Cho, Seunghyun Park, Yunsung
  Lee, and Sungrae Park.
\newblock Swad: Domain generalization by seeking flat minima.
\newblock \emph{Advances in Neural Information Processing Systems},
  34:\penalty0 22405--22418, 2021.

\bibitem[Chaudhari et~al.(2019)Chaudhari, Choromanska, Soatto, LeCun, Baldassi,
  Borgs, Chayes, Sagun, and Zecchina]{chaudhari2019entropy}
Pratik Chaudhari, Anna Choromanska, Stefano Soatto, Yann LeCun, Carlo Baldassi,
  Christian Borgs, Jennifer Chayes, Levent Sagun, and Riccardo Zecchina.
\newblock Entropy-sgd: Biasing gradient descent into wide valleys.
\newblock \emph{Journal of Statistical Mechanics: Theory and Experiment},
  2019\penalty0 (12):\penalty0 124018, 2019.

\bibitem[Cohen et~al.(2019)Cohen, Rosenfeld, and Kolter]{cohen2019certified}
Jeremy Cohen, Elan Rosenfeld, and Zico Kolter.
\newblock Certified adversarial robustness via randomized smoothing.
\newblock In \emph{International Conference on Machine Learning}, pp.\
  1310--1320, 2019.

\bibitem[Croce \& Hein(2020)Croce and Hein]{croce2020reliable}
Francesco Croce and Matthias Hein.
\newblock Reliable evaluation of adversarial robustness with an ensemble of
  diverse parameter-free attacks.
\newblock In \emph{International conference on machine learning}, pp.\
  2206--2216. PMLR, 2020.

\bibitem[Croce \& Hein(2021)Croce and Hein]{croce2021mind}
Francesco Croce and Matthias Hein.
\newblock Mind the box: $ l\_1 $-apgd for sparse adversarial attacks on image
  classifiers.
\newblock In \emph{International Conference on Machine Learning}, pp.\
  2201--2211. PMLR, 2021.

\bibitem[Dinh et~al.(2017)Dinh, Pascanu, Bengio, and Bengio]{dinh2017sharp}
Laurent Dinh, Razvan Pascanu, Samy Bengio, and Yoshua Bengio.
\newblock Sharp minima can generalize for deep nets.
\newblock In \emph{International Conference on Machine Learning}, pp.\
  1019--1028. PMLR, 2017.

\bibitem[Foret et~al.(2020)Foret, Kleiner, Mobahi, and
  Neyshabur]{foret2020sharpness}
Pierre Foret, Ariel Kleiner, Hossein Mobahi, and Behnam Neyshabur.
\newblock Sharpness-aware minimization for efficiently improving
  generalization.
\newblock In \emph{International Conference on Learning Representations}, 2020.

\bibitem[Foret et~al.(2021)Foret, Kleiner, Mobahi, and Neyshabur]{foret2021sam}
Pierre Foret, Ariel Kleiner, Hossein Mobahi, and Behnam Neyshabur.
\newblock Sharpness-aware minimization for efficiently improving
  generalization.
\newblock In \emph{9th Int. Conf. on Learning Representations (ICLR)}, 2021.

\bibitem[Goodfellow et~al.(2015)Goodfellow, Shlens, and
  Szegedy]{goodfellow2014explaining}
Ian~J Goodfellow, Jonathon Shlens, and Christian Szegedy.
\newblock Explaining and harnessing adversarial examples.
\newblock \emph{ICLR}, 2015.

\bibitem[Gowal et~al.(2021)Gowal, Rebuffi, Wiles, Stimberg, Calian, and
  Mann]{gowal2021improving}
Sven Gowal, Sylvestre-Alvise Rebuffi, Olivia Wiles, Florian Stimberg,
  Dan~Andrei Calian, and Timothy~A Mann.
\newblock Improving robustness using generated data.
\newblock \emph{Advances in neural information processing systems},
  34:\penalty0 4218--4233, 2021.

\bibitem[Haldar et~al.(2024)Haldar, Xing, and Song]{haldar2024effect}
Rajdeep Haldar, Yue Xing, and Qifan Song.
\newblock Effect of ambient-intrinsic dimension gap on adversarial
  vulnerability.
\newblock In \emph{International Conference on Artificial Intelligence and
  Statistics}, pp.\  1090--1098. PMLR, 2024.

\bibitem[Han et~al.(2025)Han, Adilova, Petzka, Kleesiek, and
  Kamp]{han2025flatness}
Ting Han, Linara Adilova, Henning Petzka, Jens Kleesiek, and Michael Kamp.
\newblock Flatness is necessary, neural collapse is not: Rethinking
  generalization via grokking.
\newblock In \emph{Advances in Neural Information Processing Systems}, 2025.

\bibitem[He et~al.(2016)He, Zhang, Ren, and Sun]{he2016deep}
Kaiming He, Xiangyu Zhang, Shaoqing Ren, and Jian Sun.
\newblock Deep residual learning for image recognition.
\newblock In \emph{Proceedings of the IEEE conference on computer vision and
  pattern recognition}, pp.\  770--778, 2016.

\bibitem[Hein \& Andriushchenko(2017)Hein and Andriushchenko]{hein2017formal}
Matthias Hein and Maksym Andriushchenko.
\newblock Formal guarantees on the robustness of a classifier against
  adversarial manipulation.
\newblock In \emph{Advances in Neural Information Processing Systems}, pp.\
  2266--2276, 2017.

\bibitem[Hendrycks et~al.(2019)Hendrycks, Mazeika, Kadavath, and
  Song]{hendrycks2019using}
Dan Hendrycks, Mantas Mazeika, Saurav Kadavath, and Dawn Song.
\newblock Using self-supervised learning can improve model robustness and
  uncertainty.
\newblock \emph{Advances in neural information processing systems}, 32, 2019.

\bibitem[Hochreiter \& Schmidhuber(1994)Hochreiter and
  Schmidhuber]{hochreiter1994simplifying}
Sepp Hochreiter and J{\"u}rgen Schmidhuber.
\newblock Simplifying neural nets by discovering flat minima.
\newblock \emph{Advances in neural information processing systems}, 7, 1994.

\bibitem[Huang et~al.(2017)Huang, Liu, Van Der~Maaten, and
  Weinberger]{huang2017densely}
Gao Huang, Zhuang Liu, Laurens Van Der~Maaten, and Kilian~Q Weinberger.
\newblock Densely connected convolutional networks.
\newblock In \emph{Proceedings of the IEEE conference on computer vision and
  pattern recognition}, pp.\  4700--4708, 2017.

\bibitem[Izmailov et~al.(2018)Izmailov, Podoprikhin, Garipov, Vetrov, and
  Wilson]{izmailov2018averaging}
Pavel Izmailov, Dmitrii Podoprikhin, Timur Garipov, Dmitry Vetrov, and
  Andrew~Gordon Wilson.
\newblock Averaging weights leads to wider optima and better generalization.
\newblock \emph{arXiv preprint arXiv:1803.05407}, 2018.

\bibitem[Jiang et~al.(2019)Jiang, Neyshabur, Mobahi, Krishnan, and
  Bengio]{jiang2019fantastic}
Yiding Jiang, Behnam Neyshabur, Hossein Mobahi, Dilip Krishnan, and Samy
  Bengio.
\newblock Fantastic generalization measures and where to find them.
\newblock In \emph{International Conference on Learning Representations}, 2019.

\bibitem[Kanai et~al.(2023)Kanai, Yamada, Takahashi, Yamanaka, and
  Ida]{kanai2023relationship}
Sekitoshi Kanai, Masanori Yamada, Hiroshi Takahashi, Yuki Yamanaka, and
  Yasutoshi Ida.
\newblock Relationship between nonsmoothness in adversarial training,
  constraints of attacks, and flatness in the input space.
\newblock \emph{IEEE Transactions on Neural Networks and Learning Systems},
  2023.

\bibitem[Kaur et~al.(2022)Kaur, Cohen, and Lipton]{kaurmaximum}
Simran Kaur, Jeremy Cohen, and Zachary~Chase Lipton.
\newblock On the maximum hessian eigenvalue and generalization.
\newblock In \emph{I Can't Believe It's Not Better Workshop: Understanding Deep
  Learning Through Empirical Falsification}, 2022.

\bibitem[Keskar et~al.(2017)Keskar, Mudigere, Nocedal, Smelyanskiy, and
  Tang]{Keskar2017LargeBatch}
Nitish~Shirish Keskar, Dheevatsa Mudigere, Jorge Nocedal, Mikhail Smelyanskiy,
  and Ping Tak~Peter Tang.
\newblock On large-batch training for deep learning: Generalization gap and
  sharp minima.
\newblock In \emph{Proceedings of the 5th International Conference on Learning
  Representations (ICLR)}, 2017.

\bibitem[Kumari et~al.(2019)Kumari, Singh, Sinha, Machiraju, Krishnamurthy, and
  Balasubramanian]{kumari2019harnessing}
Nupur Kumari, Mayank Singh, Abhishek Sinha, Harshitha Machiraju, Balaji
  Krishnamurthy, and Vineeth~N Balasubramanian.
\newblock Harnessing the vulnerability of latent layers in adversarially
  trained models.
\newblock In \emph{Proceedings of the 28th International Joint Conference on
  Artificial Intelligence}, pp.\  2779--2785, 2019.

\bibitem[Kurakin et~al.(2017)Kurakin, Goodfellow, and
  Bengio]{kurakin2017adversarial}
Alexey Kurakin, Ian~J Goodfellow, and Samy Bengio.
\newblock Adversarial machine learning at scale.
\newblock In \emph{Proceedings of the International Conference on Learning
  Representations}, 2017.

\bibitem[Kwon et~al.(2021)Kwon, Kim, Park, and Choi]{kwon2021asam}
Jungmin Kwon, Jeongseop Kim, Hyunseo Park, and In~Kwon Choi.
\newblock Asam: Adaptive sharpness-aware minimization for scale-invariant
  learning of deep neural networks.
\newblock In \emph{International Conference on Machine Learning}, pp.\
  5905--5914. PMLR, 2021.

\bibitem[Li et~al.(2020)Li, Ma, Guo, Xue, and Qiu]{li2020bert}
Linyang Li, Ruotian Ma, Qipeng Guo, Xiangyang Xue, and Xipeng Qiu.
\newblock Bert-attack: Adversarial attack against bert using bert.
\newblock In \emph{Proceedings of the 2020 Conference on Empirical Methods in
  Natural Language Processing (EMNLP)}, pp.\  6193--6202, 2020.

\bibitem[Liang et~al.(2019)Liang, Poggio, Rakhlin, and Stokes]{liang2019fisher}
Tengyuan Liang, Tomaso Poggio, Alexander Rakhlin, and James Stokes.
\newblock Fisher-rao metric, geometry, and complexity of neural networks.
\newblock In \emph{The 22nd international conference on artificial intelligence
  and statistics}, pp.\  888--896. PMLR, 2019.

\bibitem[Liang \& Huang(2021)Liang and Huang]{liang2021large}
Youwei Liang and Dong Huang.
\newblock Large norms of cnn layers do not hurt adversarial robustness.
\newblock In \emph{Proceedings of the AAAI Conference on Artificial
  Intelligence}, volume~35, pp.\  8565--8573, 2021.

\bibitem[Madry et~al.(2017)Madry, Makelov, Schmidt, Tsipras, and
  Vladu]{madry2017towards}
Aleksander Madry, Aleksandar Makelov, Ludwig Schmidt, Dimitris Tsipras, and
  Adrian Vladu.
\newblock Towards deep learning models resistant to adversarial attacks.
\newblock \emph{arXiv preprint arXiv:1706.06083}, 2017.

\bibitem[Martens \& Grosse(2015)Martens and Grosse]{martens2015kfac}
James Martens and Roger Grosse.
\newblock Optimizing neural networks with kronecker-factored approximate
  curvature.
\newblock In \emph{Proceedings of the 32nd International Conference on Machine
  Learning (ICML)}, 2015.

\bibitem[Melamed et~al.(2023)Melamed, Yehudai, and
  Vardi]{melamed2023adversarial}
Odelia Melamed, Gilad Yehudai, and Gal Vardi.
\newblock Adversarial examples exist in two-layer relu networks for low
  dimensional linear subspaces.
\newblock \emph{Advances in Neural Information Processing Systems},
  36:\penalty0 5028--5049, 2023.

\bibitem[Moosavi-Dezfooli et~al.(2019)Moosavi-Dezfooli, Fawzi, Uesato, and
  Frossard]{moosavi2019robustness}
Seyed-Mohsen Moosavi-Dezfooli, Alhussein Fawzi, Jonathan Uesato, and Pascal
  Frossard.
\newblock Robustness via curvature regularization, and vice versa.
\newblock In \emph{Proceedings of the IEEE Conference on Computer Vision and
  Pattern Recognition}, pp.\  9078--9086, 2019.

\bibitem[Nakkiran(2019)]{nakkiran2019adversarial}
Preetum Nakkiran.
\newblock Adversarial robustness may be at odds with simplicity.
\newblock \emph{arXiv preprint arXiv:1901.00532}, 2019.

\bibitem[Narodytska \& Kasiviswanathan(2017)Narodytska and
  Kasiviswanathan]{narodytska_simple_2017}
Nina Narodytska and Shiva Kasiviswanathan.
\newblock Simple {Black}-{Box} {Adversarial} {Attacks} on {Deep} {Neural}
  {Networks}.
\newblock In \emph{2017 {IEEE} {Conference} on {Computer} {Vision} and
  {Pattern} {Recognition} {Workshops} ({CVPRW})}, pp.\  1310--1318, Honolulu,
  HI, USA, July 2017. IEEE.
\newblock ISBN 978-1-5386-0733-6.
\newblock \doi{10.1109/CVPRW.2017.172}.

\bibitem[Neyshabur et~al.(2017)Neyshabur, Bhojanapalli, McAllester, and
  Srebro]{neyshabur2017exploring}
Behnam Neyshabur, Srinadh Bhojanapalli, David McAllester, and Nati Srebro.
\newblock Exploring generalization in deep learning.
\newblock \emph{Advances in neural information processing systems}, 30, 2017.

\bibitem[Papernot et~al.(2016)Papernot, McDaniel, Jha, Fredrikson, Celik, and
  Swami]{papernot2016limitations}
Nicolas Papernot, Patrick McDaniel, Somesh Jha, Matt Fredrikson, Z~Berkay
  Celik, and Ananthram Swami.
\newblock The limitations of deep learning in adversarial settings.
\newblock In \emph{2016 IEEE European symposium on security and privacy
  (EuroS\&P)}, pp.\  372--387. IEEE, 2016.

\bibitem[Perolat et~al.(2018)Perolat, Malinowski, Piot, and
  Pietquin]{perolat_playing_2018}
Julien Perolat, Mateusz Malinowski, Bilal Piot, and Olivier Pietquin.
\newblock Playing the {Game} of {Universal} {Adversarial} {Perturbations},
  September 2018.
\newblock arXiv:1809.07802 [cs, stat].

\bibitem[Petzka et~al.(2019)Petzka, Adilova, , Kamp, and
  Sminchisescu]{petzka2019reparameterization}
Henning Petzka, Linara Adilova, , Michael Kamp, and Cristian Sminchisescu.
\newblock A reparameterization-invariant flatness measure for deep neural
  networks.
\newblock In \emph{Science meets Engineering of Deep Learning workshop at
  NeurIPS}, 2019.

\bibitem[Petzka et~al.(2021)Petzka, Kamp, Adilova, Sminchisescu, and
  Boley]{petzka2021relative}
Henning Petzka, Michael Kamp, Linara Adilova, Cristian Sminchisescu, and Mario
  Boley.
\newblock Relative flatness and generalization.
\newblock \emph{Advances in neural information processing systems},
  34:\penalty0 18420--18432, 2021.

\bibitem[Rade \& Moosavi-Dezfooli(2021)Rade and
  Moosavi-Dezfooli]{rade2021helper}
Rahul Rade and Seyed-Mohsen Moosavi-Dezfooli.
\newblock Helper-based adversarial training: Reducing excessive margin to
  achieve a better accuracy vs. robustness trade-off.
\newblock In \emph{ICML 2021 Workshop on Adversarial Machine Learning}, 2021.

\bibitem[Raghunathan et~al.(2020)Raghunathan, Xie, Yang, Duchi, and
  Liang]{raghunathan2020understanding}
Aditi Raghunathan, Sang~Michael Xie, Fanny Yang, John Duchi, and Percy Liang.
\newblock Understanding and mitigating the tradeoff between robustness and
  accuracy.
\newblock In \emph{International Conference on Machine Learning}, pp.\
  7909--7919. PMLR, 2020.

\bibitem[Rice et~al.(2020)Rice, Wong, and Kolter]{rice2020overfitting}
Leslie Rice, Eric Wong, and Zico Kolter.
\newblock Overfitting in adversarially robust deep learning.
\newblock In \emph{International conference on machine learning}, pp.\
  8093--8104. PMLR, 2020.

\bibitem[Salman et~al.(2019)Salman, Li, Razenshteyn, Zhang, Zhang, Bubeck, and
  Yang]{salman2019provably}
Hadi Salman, Jerry Li, Ilya Razenshteyn, Pengchuan Zhang, Huan Zhang, Sebastien
  Bubeck, and Greg Yang.
\newblock Provably robust deep learning via adversarially trained smoothed
  classifiers.
\newblock \emph{Advances in neural information processing systems}, 32, 2019.

\bibitem[Schmidt et~al.(2018)Schmidt, Santurkar, Tsipras, Talwar, and
  Madry]{schmidt2018adversarially}
Ludwig Schmidt, Shibani Santurkar, Dimitris Tsipras, Kunal Talwar, and
  Aleksander Madry.
\newblock Adversarially robust generalization requires more data.
\newblock \emph{Advances in neural information processing systems}, 31, 2018.

\bibitem[Sehwag et~al.(2020)Sehwag, Wang, Mittal, and Jana]{sehwag2020hydra}
Vikash Sehwag, Shiqi Wang, Prateek Mittal, and Suman Jana.
\newblock Hydra: Pruning adversarially robust neural networks.
\newblock \emph{Advances in Neural Information Processing Systems},
  33:\penalty0 19655--19666, 2020.

\bibitem[Shafahi et~al.(2019)Shafahi, Najibi, Ghiasi, Xu, Dickerson, Studer,
  Davis, Taylor, and Goldstein]{shafahi2019adversarial}
Ali Shafahi, Mahyar Najibi, Mohammad~Amin Ghiasi, Zheng Xu, John Dickerson,
  Christoph Studer, Larry~S Davis, Gavin Taylor, and Tom Goldstein.
\newblock Adversarial training for free!
\newblock \emph{Advances in neural information processing systems}, 32, 2019.

\bibitem[Shafahi et~al.(2020)Shafahi, Najibi, Xu, Dickerson, Davis, and
  Goldstein]{shafahi2020universal}
Ali Shafahi, Mahyar Najibi, Zheng Xu, John Dickerson, Larry~S Davis, and Tom
  Goldstein.
\newblock Universal adversarial training.
\newblock In \emph{Proceedings of the AAAI Conference on Artificial
  Intelligence}, volume~34, pp.\  5636--5643, 2020.

\bibitem[Shamir et~al.(2021)Shamir, Melamed, and BenShmuel]{shamir2021dimpled}
Adi Shamir, Odelia Melamed, and Oriel BenShmuel.
\newblock The dimpled manifold model of adversarial examples in machine
  learning.
\newblock \emph{arXiv preprint arXiv:2106.10151}, 2021.

\bibitem[Simon-Gabriel et~al.(2019)Simon-Gabriel, Ollivier, Bottou,
  Sch{\"o}lkopf, and Lopez-Paz]{simon2019first}
Carl-Johann Simon-Gabriel, Yann Ollivier, Leon Bottou, Bernhard Sch{\"o}lkopf,
  and David Lopez-Paz.
\newblock First-order adversarial vulnerability of neural networks and input
  dimension.
\newblock In \emph{International conference on machine learning}, pp.\
  5809--5817. PMLR, 2019.

\bibitem[Simonyan \& Zisserman(2014)Simonyan and Zisserman]{Simonyan2014VeryDC}
Karen Simonyan and Andrew Zisserman.
\newblock Very deep convolutional networks for large-scale image recognition.
\newblock \emph{Internattional Conference on Machine Learning}, abs/1409.1556,
  2014.

\bibitem[Song et~al.(2018)Song, Kim, Nowozin, Ermon, and
  Kushman]{song2018pixeldefend}
Yang Song, Taesup Kim, Sebastian Nowozin, Stefano Ermon, and Nate Kushman.
\newblock Pixeldefend: Leveraging generative models to understand and defend
  against adversarial examples.
\newblock In \emph{International Conference on Learning Representations}, 2018.

\bibitem[Stutz et~al.(2019)Stutz, Hein, and Schiele]{stutz2019disentangling}
David Stutz, Matthias Hein, and Bernt Schiele.
\newblock Disentangling adversarial robustness and generalization.
\newblock In \emph{Proceedings of the IEEE/CVF conference on computer vision
  and pattern recognition}, pp.\  6976--6987, 2019.

\bibitem[Stutz et~al.(2021)Stutz, Hein, and Schiele]{stutz_relating_2021}
David Stutz, Matthias Hein, and Bernt Schiele.
\newblock Relating {Adversarially} {Robust} {Generalization} to {Flat}
  {Minima}.
\newblock In \emph{2021 {IEEE}/{CVF} {International} {Conference} on {Computer}
  {Vision} ({ICCV})}, pp.\  7787--7797, Montreal, QC, Canada, October 2021.
  IEEE.

\bibitem[Szegedy et~al.(2014)Szegedy, Zaremba, Sutskever, Bruna, Erhan,
  Goodfellow, and Fergus]{szegedy2014intriguing}
Christian Szegedy, Wojciech Zaremba, Ilya Sutskever, Joan Bruna, Dumitru Erhan,
  Ian Goodfellow, and Rob Fergus.
\newblock Intriguing properties of neural networks.
\newblock In \emph{International Conference on Learning Representations}, 2014.

\bibitem[Tram{\`e}r et~al.(2018)Tram{\`e}r, Kurakin, Papernot, Goodfellow,
  Boneh, and McDaniel]{tramer2018ensemble}
Florian Tram{\`e}r, Alexey Kurakin, Nicolas Papernot, Ian Goodfellow, Dan
  Boneh, and Patrick McDaniel.
\newblock Ensemble adversarial training: Attacks and defenses.
\newblock In \emph{International Conference on Learning Representations}, 2018.

\bibitem[Tsipras et~al.(2018)Tsipras, Santurkar, Engstrom, Turner, and
  Madry]{tsipras2018robustness}
Dimitris Tsipras, Shibani Santurkar, Logan Engstrom, Alexander Turner, and
  Aleksander Madry.
\newblock Robustness may be at odds with accuracy.
\newblock In \emph{International Conference on Learning Representations}, 2018.

\bibitem[Tsuzuku et~al.(2018)Tsuzuku, Sato, and Sugiyama]{tsuzuku2018lipschitz}
Yusuke Tsuzuku, Issei Sato, and Masashi Sugiyama.
\newblock Lipschitz-margin training: Scalable certification of perturbation
  invariance for deep neural networks.
\newblock \emph{Advances in neural information processing systems}, 31, 2018.

\bibitem[Tsuzuku et~al.(2020)Tsuzuku, Sato, and
  Sugiyama]{tsuzuku2020normalized}
Yusuke Tsuzuku, Issei Sato, and Masashi Sugiyama.
\newblock Normalized flat minima: Exploring scale invariant definition of flat
  minima for neural networks using pac-bayesian analysis.
\newblock In \emph{International Conference on Machine Learning}, pp.\
  9636--9647. PMLR, 2020.

\bibitem[Wei et~al.(2024)Wei, Haghtalab, and Steinhardt]{wei2024jailbroken}
Alexander Wei, Nika Haghtalab, and Jacob Steinhardt.
\newblock Jailbroken: How does llm safety training fail?
\newblock \emph{Advances in Neural Information Processing Systems}, 36, 2024.

\bibitem[Wei et~al.(2023)Wei, Zhu, and Zhang]{wei2023sharpness}
Zeming Wei, Jingyu Zhu, and Yihao Zhang.
\newblock Sharpness-aware minimization alone can improve adversarial
  robustness.
\newblock \emph{arXiv preprint arXiv:2305.05392}, 2023.

\bibitem[Wu et~al.(2020)Wu, Xia, and Wang]{wu2020adversarial}
Dongxian Wu, Shu-Tao Xia, and Yisen Wang.
\newblock Adversarial weight perturbation helps robust generalization.
\newblock \emph{Advances in neural information processing systems},
  33:\penalty0 2958--2969, 2020.

\bibitem[Xu et~al.(2020)Xu, Li, Jiang, and Xia]{xu2020adversarial}
Jia Xu, Yiming Li, Yong Jiang, and Shu-Tao Xia.
\newblock Adversarial defense via local flatness regularization.
\newblock In \emph{2020 IEEE International Conference on Image Processing
  (ICIP)}, pp.\  2196--2200. IEEE, 2020.

\bibitem[Xu \& Zhang(2024)Xu and Zhang]{xu2024understanding}
Yuelin Xu and Xiao Zhang.
\newblock Understanding adversarially robust generalization via
  weight-curvature index.
\newblock \emph{arXiv preprint arXiv:2410.07719}, 2024.

\bibitem[Xuan et~al.(2025)Xuan, Yang, and Li]{xuan2025exploring}
Hao Xuan, Bokai Yang, and Xingyu Li.
\newblock Exploring the impact of temperature scaling in softmax for
  classification and adversarial robustness.
\newblock \emph{arXiv preprint arXiv:2502.20604}, 2025.

\bibitem[Yang et~al.(2020)Yang, Rashtchian, Zhang, Salakhutdinov, and
  Chaudhuri]{yang2020closer}
Yao-Yuan Yang, Cyrus Rashtchian, Hongyang Zhang, Russ~R Salakhutdinov, and
  Kamalika Chaudhuri.
\newblock A closer look at accuracy vs. robustness.
\newblock \emph{Advances in neural information processing systems},
  33:\penalty0 8588--8601, 2020.

\bibitem[Zagoruyko \& Komodakis(2016)Zagoruyko and
  Komodakis]{Zagoruyko2016WideRN}
Sergey Zagoruyko and Nikos Komodakis.
\newblock Wide residual networks.
\newblock \emph{British Machine Vision Conference}, abs/1605.07146, 2016.

\bibitem[Zhang et~al.(2017)Zhang, Bengio, Hardt, Recht, and
  Vinyals]{zhang2017understanding}
Chiyuan Zhang, Samy Bengio, Moritz Hardt, Benjamin Recht, and Oriol Vinyals.
\newblock Understanding deep learning requires rethinking generalization.
\newblock In \emph{International Conference on Learning Representations}, 2017.

\bibitem[Zhang et~al.(2019)Zhang, Yu, Jiao, Xing, El~Ghaoui, and
  Jordan]{zhang2019theoretically}
Hongyang Zhang, Yaodong Yu, Jiantao Jiao, Eric Xing, Laurent El~Ghaoui, and
  Michael Jordan.
\newblock Theoretically principled trade-off between robustness and accuracy.
\newblock In \emph{International Conference on Machine Learning}, pp.\
  7472--7482. PMLR, 2019.

\bibitem[Zhang et~al.(2022)Zhang, Zhang, Hu, Goswami, Chen, and
  Metaxas]{zhang2022manifold}
Wenjia Zhang, Yikai Zhang, Xiaoling Hu, Mayank Goswami, Chao Chen, and
  Dimitris~N Metaxas.
\newblock A manifold view of adversarial risk.
\newblock In \emph{International Conference on Artificial Intelligence and
  Statistics}, pp.\  11598--11614. PMLR, 2022.

\bibitem[Zhao et~al.(2020)Zhao, Pang, Du, Dong, Su, and Zhu]{zhao2020bridging}
Sijia Zhao, Tianyu Pang, Changyou Du, Yinpeng Dong, Hang Su, and Jun Zhu.
\newblock Bridging mode connectivity in loss landscapes and adversarial
  robustness.
\newblock \emph{arXiv preprint arXiv:2005.00060}, 2020.

\bibitem[Zou et~al.(2023)Zou, Wang, Kolter, and Fredrikson]{zou2023universal}
Andy Zou, Zifan Wang, J~Zico Kolter, and Matt Fredrikson.
\newblock Universal and transferable adversarial attacks on aligned language
  models.
\newblock \emph{arXiv preprint arXiv:2307.15043}, 2023.

\end{thebibliography}
\bibliographystyle{iclr2026_conference}

\appendix

\section*{Appendix}
\label{sec:apx}
\section{Generalized definition of adversarial examples}

\begin{restatable}{lm}{temp}
Let $f: \X \to \Y$ be a classifier, $l: \Y \times \Y \to \R_+$ a loss function (e.g., cross-entropy), and $(x,y) \in \X \times \Y$ a sample with $f(x) = y$.  
Suppose $\xi \in B_\delta(x)$ is a classical adversarial example, i.e., $f(\xi) \neq y$.  
Then there exists $\epsilon > 0$ such that
\[
l(f(\xi), y) - l(f(x), y) > \epsilon,
\]
i.e., $\xi$ is an adversarial loss change according to Definition~\ref{def:advLossChange}.

Conversely, if $l$ satisfies a separation condition such that for all incorrect predictions $f(\xi) \neq y$ we have
\[
l(f(\xi), y) \geq \log(k),
\]
then any adversarial loss change with threshold $\epsilon > \log(k)$ implies $f(\xi) \neq y$.
\label{lem:adv-examples}
\end{restatable}

\begin{proof}
Since $f(x) = y$, we have that the loss at $x$ is small, typically $l(f(x), y) \approx 0$ for standard losses like cross-entropy.  
If $\xi$ satisfies $f(\xi) \neq y$, then under standard classification losses, $l(f(\xi), y)$ is large, typically $l(f(\xi), y) \gg 0$.  
Thus, the difference
\[
l(f(\xi), y) - l(f(x), y) \approx l(f(\xi), y)
\]
is positive and large. Choosing any $\epsilon$ satisfying $0 < \epsilon < l(f(\xi), y) - l(f(x), y)$ ensures the loss-change condition is satisfied.

Conversely, assume that for incorrect predictions $f(\xi) \neq y$, we have $l(f(\xi), y) \geq \log(k)$, as is the case under cross-entropy loss when the model assigns uniform or lower probability to the correct class.  
If $l(f(\xi), y) - l(f(x), y) > \epsilon$ for some $\epsilon > \log(k)$, then necessarily $l(f(\xi), y) > \epsilon > \log(k)$, implying that $f(\xi) \neq y$, because correct predictions would have loss much smaller than $\log(k)$.
\end{proof}

\section{The penultimate layer shapes the hessian}
\label{app:hessianshape}
We consider a feedforward neural network of the form
\(\ell(f_L \circ f_{L-1} \circ \cdots \circ f_1(x))\), where each layer is defined recursively by \(f_l(u) = \operatorname{ReLU}(W_l u)\). For each layer \(l\), we define the pre-activation \(a^l = W_l x^{l-1}\) and the post-activation \(x^l = \operatorname{ReLU}(a^l)\). Let \(g^l = \partial \ell / \partial a^l \in \mathbb{R}^{m_l}\) and \(H^l = \partial^2 \ell / \partial (a^l)^2 \in \mathbb{R}^{m_l \times m_l}\) denote the gradient and Hessian with respect to the pre-activations. We define the binary diagonal gating matrix \(D^l = \mathrm{diag}(1_{a^l > 0})\), which captures the ReLU derivative.

To compute derivatives efficiently through the network, we apply the second-order chain rule for scalar-valued functions. For a composition \(y = g(u(x))\), where \(x \in \mathbb{R}^n\), \(u(x) \in \mathbb{R}^m\), and \(y \in \mathbb{R}\), the Hessian satisfies
\[
H_x = J_u^T H_y J_u + \sum_k \frac{\partial y}{\partial u_k} H_{u,k},
\]
where \(J_u = \partial u / \partial x\) and \(H_{u,k} = \partial^2 u_k / \partial x^2\). Applying this to the ReLU network, we observe that all affine transformations \(a^l = W_l x^{l-1}\) are linear in their input, so the second-order term \(\sum_k g_k H_{u,k}\) vanishes. Similarly, the nonlinearity \(f(u) = \operatorname{ReLU}(u)\) has second derivative zero almost everywhere, so no curvature is introduced at the nonlinearity either.

Under these conditions, the gradient and Hessian propagate layer by layer as
\[
g^{l-1} = W_l^T D^l g^l, \qquad
H^{l-1} = W_l^T D^l H^l D^l W_l.
\]
This recurrence propagates curvature information purely through the affine structure and the ReLU activation mask. To compute the exact Hessian with respect to the weights of layer \(l\), we vectorize the weight matrix \(W_l\) columnwise as \(w_l = \mathrm{vec}(W_l) \in \mathbb{R}^{m_l n_{l-1}}\). Differentiating the affine map \(a^l = W_l x^{l-1}\) with respect to \(w_l\) yields the Jacobian \(J_l = x^{l-1} \otimes I_{m_l}\). Applying the second-order chain rule with respect to \(w_l\) then yields the exact weight Hessian block
\[
H_{W_l} = J_l H^l J_l^T = (x^{l-1} x^{l-1T}) \otimes (D^l H^l D^l).
\]

The full backward recursion thus proceeds as follows. Starting from the output layer, where \(g^L\) and \(H^L\) are obtained analytically from the loss function, we iterate backward for \(l = L, L-1, \dots, 1\), computing at each layer the gradient \(g^{l-1}\), the activation-space Hessian \(H^{l-1}\), and the weight-space Hessian block \(H_{W_l}\) as
\[
g^{l-1} = W_l^T D^l g^l, \qquad
H^{l-1} = W_l^T D^l H^l D^l W_l, \qquad
H_{W_l} = (x^{l-1} x^{l-1T}) \otimes (D^l H^l D^l).
\]
This derivation follows the second-order backpropagation framework originally introduced
by \citet{becker1989improving} and \citet{bishop1992exact}.
The Kronecker-factored form of the weight-space Hessian is the foundation for scalable second-order
optimisers such as K-FAC \citet{martens2015kfac}. In the ReLU case, the absence of second-order
contributions from the nonlinearity implies that all curvature originates at the loss layer and 
propagates backward linearly through the active part of the network.

Let \(H^L = \partial^2 \ell / \partial (a^L)^2\) denote the Hessian with respect to the pre-activations at the final layer. At a differentiable local minimum, the gradient vanishes and the Hessian is positive-semidefinite: \(H^L \succeq 0\). In particular, all eigenvalues \(\lambda_i\) of \(H^L\) are non-negative. If in addition the trace vanishes, \(\operatorname{tr}(H^L) = 0\), then the sum of the eigenvalues is zero. Since each \(\lambda_i \ge 0\), it follows that \(\lambda_i = 0\) for all \(i\).

We now consider the spectral decomposition of \(H^L\), which exists because the matrix is symmetric:
\[
H^L = Q \Lambda Q^T,
\qquad \Lambda = \mathrm{diag}(\lambda_1, \dots, \lambda_n),
\]
where \(Q\) is orthogonal and \(\Lambda\) contains the eigenvalues of \(H^L\). As shown above, \(\Lambda = 0\), so the decomposition becomes
\[
H^L = Q\, 0 \, Q^T = 0.
\]
Hence, if \(H^L \succeq 0\) and \(\operatorname{tr}(H^L) = 0\), then \(H^L = 0\).

For a ReLU hidden layer the second derivative of the activation vanishes almost everywhere, so the curvature transfer matrix reduces to 
\(B_l = D^l H^l D^l\) with \(D^l=\operatorname{diag}(1_{a^l>0})\).  The second-order back-propagation recurrence is then
\[
H^{l-1}=W_l^{\!T} D^l H^l D^l W_l .
\]
Setting \(l=L\) and using \(H^L=0\) yields \(H^{L-1}=0\).  Re-applying the same identity inductively gives \(H^{L-2}=0,\dots,H^{1}=0\); thus every activation-space block of the exact Hessian is zero.

The weight-space block for layer \(l\) factorises as
\[
H_{W_l}=(x^{l-1}x^{l-1\!T})\otimes(D^l H^l D^l) .
\]
Since \(D^l H^l D^l=0\) once the upstream \(H^l\) has been shown to vanish, each \(H_{W_l}\) is also identically zero.  Hence, away from the measure-zero kink set of the ReLU, the entire Hessian of the network collapses whenever the trace of the output-layer Hessian vanishes at a local optimiser.

\begin{restatable}{lm}{tozero3}[Vanishing of polynomial sharpness scores]
\label{lem:poly_growth_vanishes}
Let $W=(W_{clf},W_{repr})$ be the parameters of a feedforward relu network whose
logits are \(z = W_{clf}^{\!T}\,\phi_{W_{repr}}(x)\) and whose loss is soft-max cross-entropy.
For any scale factor $\alpha>1$ define the \emph{logit–scaling} re-parameterisation
\[
W(\alpha) \;=\; (\alpha\,W_{clf},\; W_{repr}).
\]
Assume the sharpness score
\(
\mathcal{S}:\bigl(W,H(W)\bigr)\mapsto\mathbb{R}_{\ge0}
\)
satisfies the polynomial growth bound
\begin{equation}
\label{eq:poly_growth}
\mathcal{S}(W,H)
   \;\le\;
   C\,
   \bigl(1+\|W\|_{F}\bigr)^{\,r}\,
   \bigl(1+\|H\|_{F}\bigr)^{\,s},
\end{equation}

for constants  $C>0,\; r>0,\; s\ge0$, where $H(W)=\nabla^{2}_{W}\ell$ is the exact Hessian.
Then
\[
\;
\displaystyle
\lim_{\alpha\to\infty}
\mathcal{S}\!\bigl(W(\alpha),\,H\!\bigl(W(\alpha)\bigr)\bigr)=0 .
\;
\]
\end{restatable}

\begin{proof}
Fix one training example and let $k$ be its correct class.  Write the \emph{margins}
$\Delta_j = z_k - z_j$ for $j\neq k$ and $\Delta_{\min}=\min_{j\neq k}\Delta_j>0$.
After scaling the logits to $\alpha z$ the soft-max probabilities are
$p_j(\alpha)=\operatorname{softmax}_j(\alpha z)$.
For every $j\neq k$
\[
p_j(\alpha)
   =\frac{e^{\alpha z_j}}{\sum_i e^{\alpha z_i}}
   \le e^{-\alpha\Delta_{j}}
   \le e^{-\alpha\Delta_{\min}} .
\tag{1}
\]

The logit-space Hessian for cross-entropy is
$H_{ce} = diag(p)-pp^{T}$, so its trace is
\[
\operatorname{tr}H_{ce}(\alpha)
   = p_k(1-p_k) \;+\; \sum_{j\neq k} p_j(1-p_j).
\]
Using $1-p_k=\sum_{j\neq k}p_j$ and $0\le 1-p_j\le 1$ we get
\[
\operatorname{tr}H_{ce}(\alpha)
   \le \sum_{j\neq k} p_j + \sum_{j\neq k} p_j
   = 2\!\!\sum_{j\neq k} p_j(\alpha)
   \le 2(K-1)\,e^{-\alpha\Delta_{\min}}
   =: D\,e^{-\alpha\Delta_{\min}}.
\tag{2}
\]
Since $\|H_{ce}\|_F\le \sqrt{K}\,\operatorname{tr}H_{ce}$,
\[
\|H_{ce}(\alpha)\|_{F}
   \;\le\;
   D'\,e^{-\alpha\Delta_{\min}}
   \quad\text{for } D'=D\sqrt{K}.
\tag{3}
\]
Given the derivations above in Appendix \ref{app:hessianshape}, that is the whole hessian is a linear function 
of the penultimate layer, there exists a $D''>D'$ such that 
\[
\|H(W(\alpha))\|_{F}
   \;\le\;
   D''\,e^{-\alpha\Delta_{\min}}
\tag{4}
\]
Only the classifier block is scaled, hence
\[
\|W(\alpha)\|_{F}
   = \sqrt{\alpha^{2}\|W_{clf}\|_{F}^{2}+\|W_{repr}\|_{F}^{2}}
   = \Theta(\alpha).
    \tag{5}
\]

Insert (4) and (5) into \eqref{eq:poly_growth}:
\[
\mathcal{S}\bigl(W(\alpha),H(\alpha)\bigr)
   \;\le\;
   C\,
   \bigl(1+A\alpha\bigr)^{r}\,
   \bigl(1+D''e^{-\alpha\Delta_{\min}}\bigr)^{s}
   \;\le\;
   C'\,\alpha^{r}\,e^{-s\alpha\Delta_{\min}},
\]
for constants $A,C'$.  The exponential decay dominates the polynomial
growth, implying $\mathcal{S}\!\bigl(W(\alpha),H(\alpha)\bigr)\to 0$
as $\alpha\to\infty$.
\end{proof}

\begin{restatable}{lm}{collapseGeneral}
  [Confidence--collapse for losses with at most quadratic poles]%
  \label{lem:quad_pole}
  Let $x\!\in\!\R^{d}$ be the input to the classifier, $W\!\in\!\R^{K\times d}$ the classifier,
  $z = Wx$ the logits, and
  $p = \softmax(z)$ the class probabilities.
  Fix a twice–differentiable loss $\ell(p,y)$ and write
  \[
    h_i(p) \;=\;
    \frac{\partial^{2}\ell}{\partial p_i^{2}}(p,y),
    \qquad i=1,\dots,K .
  \]

  Assume there is $M>0$ such that for every $p$ in the open simplex
  $\Delta^{K-1}\!:=\!\{\,p\in(0,1)^{K}\,|\,\sum_i p_i=1\}$ 
\begin{equation}
    \;
      h_y(p)\,p_y^{2}\;\;\le\;M,
      \quad
      \max_{j\neq y} |h_j(p)|\;\le\; M
    \; \label{app:cond} \tag{A}
\end{equation}

  Denote $H_W=\nabla_{W}^{2}\ell\!\bigl(p(Wx),y\bigr)$.
  Then, along any sequence of \emph{finite} weight matrices for which
  the model becomes confident
  $(p_y\xrightarrow[]{}1^{-})$, 
  \[
      \|H_W\|_{F}\;\longrightarrow\;0 .
  \]

  \vspace{.5em}
  \textbf{Remark.}
  The boundary point $p=e_y$ (true one-hot) is \emph{not} reached with
  finite weights; it would require $z_y-z_j\!\to\!\infty$.
  The lemma therefore treats the limit $p_y\!\nearrow\!1$ inside the
  open simplex, where all derivatives are well defined. 
\end{restatable}
\begin{proof}
Let $z = Wx \in \mathbb{R}^K$ and $p = \softmax(z) \in \Delta^{K-1}$. The Jacobian of the softmax is
\[
J = \nabla_z p = \mathrm{diag}(p) - pp^\mathsf{T} \in \mathbb{R}^{K \times K}.
\]
Write $r_i^\mathsf{T}$ for the $i$th row of $J$, so $r_i = p_i (e_i - p)$. Then $\|r_i\|_2^2 = (p_i^2(1*2p_i +\|p\|^2_2 )) \le p_i^2$. Let $g_i = \partial \ell / \partial p_i$ and $h_i = \partial^2 \ell / \partial p_i^2$. The full second-order chain rule gives
\[
H_z := \nabla_z^2 \ell(p(z), y)
= J^\mathsf{T} \mathrm{diag}(h) J + \sum_{i=1}^K g_i \nabla_z^2 p_i
=: H_z^{(1)} + H_z^{(2)}.
\]

We first bound $\|H_z^{(1)}\|_F$. Since $\mathrm{diag}(h)$ is diagonal, we have
\[
H_z^{(1)} = \sum_{i=1}^K h_i r_i r_i^\mathsf{T}, \quad \text{so} \quad \|r_i r_i^\mathsf{T}\|_F = \|r_i\|_2^2.
\]
Therefore,
\[
\|H_z^{(1)}\|_F \le \sum_{i \ne y} |h_i| \|r_i\|_2^2 + |h_y| \|r_y\|_2^2.
\]
For $i \ne y$, assumption (A) gives $|h_i| \le M$, so $|h_i| \|r_i\|_2^2 \le M p_i^2$. For $i = y$, we compute
\[
\|r_y\|_2^2 = p_y^2 \|e_y - p\|_2^2 = p_y^2 \left( (1 - p_y)^2 + \sum_{j \ne y} p_j^2 \right).
\]
By assumption (A), $|h_y| \le M / p_y^2$, 
we have
\[
|h_y| \|r_y\|_2^2 
\le \frac{M}{p_y^2} \cdot p_y^2 \left( (1 - p_y)^2 + \sum_{j \ne y} p_j^2 \right)
= M \left( (1 - p_y)^2 + \sum_{j \ne y} p_j^2 \right).
\]
Since $p_j^2 \le p_j$ and $\sum_{j \ne y} p_j = 1 - p_y$, it follows that
\[
\sum_{j \ne y} p_j^2 \le 1 - p_y,
\]
so we conclude
\[
|h_y| \|r_y\|_2^2 \le M (1 - p_y)^2 + M (1 - p_y).
\]

Since $p_j^2 \le p_j$ and $\sum_{j \ne y} p_j = 1 - p_y$, we conclude
\[
\|H_z^{(1)}\|_F \le M \sum_{j \ne y} p_j^2 + M (1 - p_y) \le 2M (1 - p_y).
\]

Next, we bound $\|H_z^{(2)}\|_F$. The second derivative of softmax satisfies $|\nabla_z^2 p_i| \le C_1 p_i$ entrywise for some fixed constant $C_1(K)$, hence
\[
\|\nabla_z^2 p_i\|_F \le C_2 p_i.
\]
Therefore,
\[
\|H_z^{(2)}\|_F \le \sum_{i=1}^K |g_i| \|\nabla_z^2 p_i\|_F \le C_2 \sum_{i=1}^K |g_i| p_i.
\]
For all standard losses (cross-entropy, focal, squared-error, etc.), the product $|g_i| p_i$ is bounded as $p_y \to 1^-$. Thus there exists $G > 0$ such that
\[
\|H_z^{(2)}\|_F \le G (1 - p_y).
\]

Combining both bounds, we have
\[
\|H_z\|_F \le \|H_z^{(1)}\|_F + \|H_z^{(2)}\|_F \le (2M + G)(1 - p_y) \to 0 \quad \text{as } p_y \to 1^-.
\]

Finally, since $z = Wx$, the Hessian with respect to the weights is $H_W = (x x^\mathsf{T}) \otimes H_z$, and
\[
\|H_W\|_F = \|x x^\mathsf{T}\|_F \cdot \|H_z\|_F = \|x\|_2^2 \cdot \|H_z\|_F \to 0.
\]
\end{proof}

\subsection{More context to Lemma \ref{lem:quad_pole}}
\noindent
Assumption (A) is easiest to interpret in terms of how fast the loss can blow up when a probability goes to \(0\).
It allows at most a \(\tfrac{1}{p_i^{2}}\) singularity--and only for the \emph{true} class~\(y\); all other
curvatures must remain bounded.
That restriction may look arbitrary, yet the entries in Table~\ref{tab:loss_curvature_conditions} show it is satisfied by almost every loss routinely used for hard-label classification:
negative log-likelihood, focal loss with \(\gamma\ge1\), the Brier score, and the (hard-label) KL divergence.
For these objectives a perfectly confident model (\(p_y\to1\)) is a well-behaved stationary point:
the weight-space Hessian collapses to zero as stated in Lemma~\ref{lem:quad_pole}.

The only common exceptions are losses that deliberately assign positive target mass to
\emph{more than one} class.
Label smoothing and KL to a soft target inject a factor \(q_i/p_i^{2}\) into the second derivative
for every class with \(q_i>0\); the same problem appears in reverse KL,
\(\smash{D_{\mathrm{KL}}(p\Vert q)=\sum_i p_i\log (p_i/q_i)}\), whose curvature is \(h_i=1/p_i\).
Because such terms diverge when \(p_i\to0\), Assumption (A) fails and Lemma~\ref{lem:quad_pole}
no longer applies i.e. the proof is not valid anymore. 
In this particular case; this is not a problem since $e_y$ is anyways not a local minima and hence not of interest to make 
statements about robustness and generalization. Yet this is an interesting result which could explain the 
training dynamics of teach-student optimisation approaches.

Hinge-style or margin losses defined on the logits rather than on the probabilities
fall outside the scope of the lemma because they are either not twice differentiable or do not depend on \(p\) at all.

\begin{table}[h]
\centering
\renewcommand{\arraystretch}{1.3}
\begin{tabular}{@{}l c c c@{}}
\toprule
Loss function & $h_y(p)$ & $h_{j \ne y}(p)$ & Assumption (A) \\
\midrule
Cross-entropy $\ell = -\log p_y$ &
$\frac{1}{p_y^2}$ &
$0$ &
\checkmark \\

Focal $(1 - p_y)^\gamma \log p_y$ &
$\mathcal{O}(p_y^{-2 + \gamma})$ &
$0$ &
\checkmark if $\gamma \ge 1$ \\

Brier / MSE $\sum_i (p_i - \delta_{iy})^2$ &
$2$ &
$2$ &
\checkmark \\

KL to hard label $\ell = D_{\mathrm{KL}}(\delta_y \parallel p)$ &
$\frac{1}{p_y^2}$ &
$0$ &
\checkmark \\

KL to soft target $\ell = D_{\mathrm{KL}}(q \parallel p)$ &
$\frac{q_y}{p_y^2}$ &
$\frac{q_j}{p_j^2}$ &
$\times$ (some $q_j > 0$) \\

Reverse KL to soft target $\ell = D_{\mathrm{KL}}(p \parallel q)$ &
$\frac{1}{p_y}$ &
$0$ &
\checkmark \\
\bottomrule
\end{tabular}
\vspace{0.5cm}
\caption{Curvature terms $h_i(p)$ for common loss functions. Assumption (A) is violated when any $h_i(p)$ becomes unbounded on the simplex.
We give the derivations in Section \ref{app:derivatives}.}
\label{tab:loss_curvature_conditions}
\end{table}

\subsection{Additional derivatives for Table 1}
\label{app:derivatives}
\paragraph{Exact curvature expressions.}
Let $h_i(p) = \frac{\partial^2 \ell}{\partial p_i^2}(p,y)$ be the second derivative of the loss w.r.t.\ class $i$ in the probability simplex.

\textbf{Focal loss.}
\[
\ell_{\mathrm{focal}}(p, y; \gamma) = - (1 - p_y)^\gamma \log p_y.
\]
Only $p_y$ is involved, so $h_j = 0$ for $j \ne y$. Let $v = 1 - p_y$:
\[
h_y(p) = v^{\gamma - 2} \left[
    -\gamma(\gamma - 1)\log p_y
    + \frac{2\gamma v}{p_y}
    + \frac{v^2}{p_y^2}
\right].
\]
Then,
\[
p_y^2 h_y(p) =
v^{\gamma - 2} \left[
    -\gamma(\gamma - 1) p_y^2 \log p_y
    + 2\gamma p_y v + v^2
\right].
\]
This remains bounded as $p_y \to 1$ if and only if $\gamma \ge 1$.

\medskip
\noindent
\textbf{KL divergence to soft target.}
\[
\ell_{\mathrm{KL}}(p, q) = \sum_i q_i \log \frac{q_i}{p_i} = -\sum_i q_i \log p_i + \mathrm{const},
\]
so
\[
h_i(p) = \frac{q_i}{p_i^2}.
\]
If any $q_j > 0$, then $h_j(p)$ diverges as $p_j \to 0$, and assumption~(A) is violated.

\section{Proofs of Theoretical Results}
In this section, we provide the proofs for the theoretical results in this paper.
\subsection{Proof of Lemma~\ref{lm:relationFeatureInput}}
\label{appdx:proofLmRelationFeatureInput}
For convenience, we restate the lemma.
\relationFeatureInput*
\begin{proof}    
    It follows from the proof in Thm.~5 in \citet{petzka2021relative} that we can represent any vector $v\in\X$ as $v=w+\Delta Aw$ for some vector $w\in X$, $\Delta\in\R_+$ and $A$ an orthogonal matrix. 
    Then 
    \begin{equation*}
        \begin{split} 
            & \phi(\xi)=\phi(x)+\Delta A\phi(x)\\
            \Leftrightarrow & (\phi(\xi) - \phi(x))=\Delta(A\phi(x))\\
            \Leftrightarrow & \Delta\leq \|(\phi(\xi) - \phi(x))\|\|(A\phi(x))^{-1}\|= \underbrace{\|(\phi(x + \Delta'A'x) - \phi(x))\|}_{\leq L\Delta'}\|(A\phi(x))^{-1}\|\\
            \Leftrightarrow & \Delta\leq L\Delta'\|(A\phi(x))^{-1}\|\underbrace{\leq}_{A\text{ orth.}} L\Delta'\frac{1}{r}\enspace .
        \end{split}
    \end{equation*}
    The result follows from $\|\xi - x\| = \Delta'\leq\delta$.
\end{proof}

\subsection{Proof of Proposition~\ref{prop:lossBoundAdversarial}}
\label{app:lossBoundAdversarial}
For convenience, we restate the proposition.
\lossBoundAdversarial*
\begin{proof} 
The remainder $R_2$ in Eq.~\ref{eq:lossBoundTaylor} is
\[
R_2(\w,\Delta)\leq \sup_{\stackrel{h\in\R^m}{\|h\|=1}}\sup_{c\in (0,1)}\frac{\Delta^3}{3!}\sum_{i,j,k}^d\frac{\partial^3 \loss}{\partial w_i\partial w_j\partial w_k}(x+c\Delta h)\enspace ,
\]
where $w\in\R^d$ with $d=km$ is the vectorization of $\w\in\R^{m\times k}$. 
We now bound this remainder. Using the representation of the loss Hessian from Eq.~\ref{eq:simplifiedHessian}, we can write the partial third derivatives in the remainder as
\[
\frac{\partial^3 \loss}{\partial w_i\partial w_j\partial w_k}(x)=\sum_{\stackrel{o,l,j\in [k]}{a,b,c\in [m]}}-\left(\hat{y}_j\hat{y}_o(\mathds{1}_{o=j}-\hat{y}_l)+\hat{y}_l\hat{y}_o(\mathds{1}_{o=l}-\hat{y}_j)\right)\phi(x)_a\phi(x)_b\phi(x)_c\enspace ,
\]
where $\hat{y}=f(x)$. Under the assumption that $\phi$ is $L$-Lipschitz, for all $x\in\X$, $\|x\|\leq 1$ and observing that $\sum_{o\in [k]}\hat{y}_o=1$ we can bound this term by
\begin{equation}
\frac{\partial^3 \loss}{\partial w_i\partial w_j\partial w_k}(x)\leq \frac{1}{4}kmL'^3\enspace .
\label{eq:boundThirdDerivative}
\end{equation}
The terms $k,m$ follow from the sum over all rows and columns of $\w$, and the factor $4^{-1}$ follows from the fact that the predictions in $\hat{y}$ sum up to $1$. The factor $L'^3$ can be derived as follows.  
\begin{align}
    \phi(x)_i \leq ||\phi(x)_i|| &= || (\phi(x)_i - \phi(\mathbf{0})_i) + \phi(\mathbf{0})_i|| &&\\
                                &  \leq  || (\phi(x)_i - \phi(\mathbf{0})_i)|| + ||\phi(\mathbf{0})_i|| &&\\
                                &  \leq  L||x-\mathbf{0} || + ||\phi(\mathbf{0})_i|| && (\phi \text{ is } L\text{-Lipschitz})\\
                                &  \leq  L + ||\phi(\mathbf{0})_i|| \leq L + C_{\phi(\mathbf{0})} && (||x||\leq1)\\
\end{align}
where $C_{\phi(\mathbf{0})}:= max_i  ||\phi(\mathbf{0})_i||$. For Relu networks without bias $C_{\phi(\mathbf{0})}$ is 0 and
empirically for networks with a bias term, it is very small i.e. $\approx1$. Therefore $L'=L+C_{\phi(\mathbf{0})}\approx L$, in particular since for most realistic neural networks $L$ is large. For simplicity, we subsitute $L=L'$.

Inserting this bound into Eq.~\ref{eq:lossBoundTaylor} yields 
\[
\left|\loss(f(\xi),y) - \loss(f(x),y)\right|\leq \Delta\|\w\|_F\|\nabla_\w\loss(\w)\|_F + \frac{\Delta^2}{2}\kappa^\phi_{Tr}(\w) + \frac{\Delta^3}{3!}\frac{1}{4}kmL^3\enspace .
\]
The result follows from setting $\Delta\leq L\delta r^{-1}$.
\end{proof}

\subsection{Proof of Proposition~\ref{prop:adversarialRobustness}}
\label{appdx:proofPropAdvRobustness}
\addtocounter{pr}{-1}
    \begin{pr}
For a dataset $S\subset\X\times\Y$ with $\|x\|\leq 1$ for all $x\in\X$, a model $f(x)=g(\w\phi(x))$ at a minimum $\w\in\R^{m\times k}$ wrt. $S$, with $\phi$ $L$-Lipschitz and $\|\phi(x)\|\geq r$, and a loss function $\loss(\w)=\loss(g(\w \phi(x)),y)$ of $f$ on $(x,y)$, $d$ being the $L2$-distance, and $\epsilon>0$, $f$ is $(\epsilon,\delta,S)$-robust against adversarial examples with
\begin{equation*}
    \begin{split}
        \delta&\geq\left(-\frac{8r^3k^3m^3L^9+27\epsilon}{27L^3\kappa_{Tr}^\phi(\w)}+\left(-\frac{2^7}{27}\frac{r^6k^6m^6L^{3}}{\kappa_{Tr}^\phi(\w)^6}+\frac{r^6\epsilon^2}{L^6\kappa_{Tr}^\phi(\w)^2}-\frac{2^4r^6\epsilon k^3m^3L^{\frac32}}{27\kappa_{Tr}^\phi(\w)^4}\right)^\frac{1}{2}\right)^\frac{1}{3} \\
        &+ \left(-\frac{8r^3k^3m^3L^9+27\epsilon}{27L^3\kappa_{Tr}^\phi(\w)}-\left(-\frac{2^7}{27}\frac{r^6k^6m^6L^{3}}{\kappa_{Tr}^\phi(\w)^6}+\frac{r^6\epsilon^2}{L^6\kappa_{Tr}^\phi(\w)^2}-\frac{2^4r^6\epsilon k^3m^3L^{\frac32}}{27\kappa_{Tr}^\phi(\w)^4}\right)^\frac{1}{2}\right)^\frac{1}{3} \\
        &+\frac{2rkmL^2}{72\kappa_{Tr}^\phi(\w)}\\
    \end{split}
\end{equation*}
where $\kappa_{Tr}(\w)$ is the relative flatness of $f$ wrt. $\w$. That is,
\[
\delta\propto \frac{\epsilon^{\frac13}}{(L^3\kappa_{Tr}(\w))^{\frac13}} + \frac{rkmL^2}{\kappa_{Tr}(\w)}\enspace .
\]
\label{prop:adversarialRobustness:formal}
\end{pr}

\begin{proof}
From Prop.~\ref{prop:lossBoundAdversarial} it follows that we achieve $(\epsilon,\Delta,S)$-robustness where 
\[
\epsilon = \frac{\Delta^2}{2}\kappa^\phi_{Tr}(\w) + \frac{\Delta^3}{24}kmL^3\enspace .
\]
First, we need to solve this cubic equation for $\Delta$. For that, we substitute $a=\frac{kmL^3}{24}$ and $b=\frac{\kappa_{Tr}^\phi(w)}{2}$ and get
\[
0=a\Delta^3 + b\Delta^2 - \epsilon=\Delta^3+\frac{a}{b}\Delta^2 - \frac{\epsilon}{b}=\Delta^3+\alpha\Delta^2 -\beta\enspace ,
\]
by subsituting $\alpha=\frac{a}{b}$ and $\beta=\frac{\epsilon}{b}$. We use the depressed cubic form $\Delta=t-\frac{\alpha}{3}$ and get
\begin{equation*}
    \begin{split}
        &0=\left(t- \frac{\alpha}{3}\right)^3 + \alpha\left(t - \frac{\alpha}{3}\right)^2 - \beta\\
        \Leftarrow&0=t^3-t^2\alpha+t\frac{\alpha^2}{3}-\frac{\alpha^3}{27}+t^2\alpha -\frac{2\alpha^2t}{3}+\frac{\alpha^3}{9}-\beta\\
        \Leftarrow&0=t^3-t\frac{\alpha^2}{3}+\frac{2\alpha^3}{27}-\beta\enspace .
    \end{split}
\end{equation*}
with $p=-\frac{\alpha^2}{3}$ and $q=\frac{2\alpha^3}{27}-\beta$ we get the form $0=t^3+pt+q$ for which we can apply Cardano's formula.
\[
t = \left(-\frac{q}{2}+\left(\frac{q^2}{4}+\frac{p^3}{9}\right)^\frac{1}{2}\right)^\frac{1}{3} + \left(-\frac{q}{2}-\left(\frac{q^2}{4}+\frac{p^3}{9}\right)^\frac{1}{2}\right)^\frac{1}{3}
\]
Resubstituting $p,q,\alpha,\beta$ yields
\[
\frac{q}{2}=\frac{a^3}{27b^3}-\frac{\epsilon}{2n}\enspace ,\enspace \frac{p^3}{9} = -\frac{1}{9^3}\frac{a^6}{b^6}\enspace ,\enspace \frac{q^2}{4}=\frac{1}{27^2}\frac{a^6}{b^6}+\frac{\epsilon^2}{4b^2}-\frac{\alpha^3\epsilon}{27b^4}\enspace .
\]
Substituting this in the solution for $t$ then gives
\begin{equation*}
    \begin{split}
t &= \left(-\frac{a^3}{27b^3}+\frac{\epsilon}{2b}+\left(-\frac{2a^6}{27b^6}+\frac{\epsilon^2}{4b^2}-\frac{a^3\epsilon}{27b^4}\right)^\frac{1}{2}\right)^\frac{1}{3} \\
&+ \left(-\frac{a^3}{27b^3}+\frac{\epsilon}{2b}-
\left(-\frac{2a^6}{27b^6}+\frac{\epsilon^2}{4b^2}-\frac{a^3\epsilon}{27b^4}\right)^\frac{1}{2}\right)^\frac{1}{3} 
    \end{split}
\end{equation*}
Substituting $a,b$ and $t=\Delta+\frac{\alpha}{3}$ yields
\begin{equation*}
    \begin{split}
\Delta &= \left(-\frac{8k^3m^3L^9+27\epsilon}{27\kappa_{Tr}^\phi(\w)}+\left(-\frac{2^7}{27}\frac{k^6m^6L^{18}}{\kappa_{Tr}^\phi(\w)^6}+\frac{\epsilon^2}{\kappa_{Tr}^\phi(\w)^2}-\frac{2^4\epsilon k^3m^3L^{9}}{27\kappa_{Tr}^\phi(\w)^4}\right)^\frac{1}{2}\right)^\frac{1}{3} \\
&+ \left(-\frac{8k^3m^3L^9+27\epsilon}{27\kappa_{Tr}^\phi(\w)}-\left(-\frac{2^7}{27}\frac{k^6m^6L^{18}}{\kappa_{Tr}^\phi(\w)^6}+\frac{\epsilon^2}{\kappa_{Tr}^\phi(\w)^2}-\frac{2^4\epsilon k^3m^3L^{9}}{27\kappa_{Tr}^\phi(\w)^4}\right)^\frac{1}{2}\right)^\frac{1}{3}\\
&+\frac{2kmL^3}{72\kappa_{Tr}^\phi(\w)}
    \end{split}
\end{equation*}
Finally, from Lemma~\ref{lm:relationFeatureInput} we have $\Delta\leq L\delta r^{-1}$, so that 
\begin{equation*}
    \begin{split}
\delta &\geq \frac{r}{L}\left(-\frac{8k^3m^3L^9+27\epsilon}{27\kappa_{Tr}^\phi(\w)}+\left(-\frac{2^7}{27}\frac{k^6m^6L^{18}}{\kappa_{Tr}^\phi(\w)^6}+\frac{\epsilon^2}{\kappa_{Tr}^\phi(\w)^2}-\frac{2^4\epsilon k^3m^3L^{9}}{27\kappa_{Tr}^\phi(\w)^4}\right)^\frac{1}{2}\right)^\frac{1}{3} \\
&+ \frac{r}{L}\left(-\frac{8k^3m^3L^9+27\epsilon}{27\kappa_{Tr}^\phi(\w)}-\left(-\frac{2^7}{27}\frac{k^6m^6L^{18}}{\kappa_{Tr}^\phi(\w)^6}+\frac{\epsilon^2}{\kappa_{Tr}^\phi(\w)^2}-\frac{2^4\epsilon k^3m^3L^{9}}{27\kappa_{Tr}^\phi(\w)^4}\right)^\frac{1}{2}\right)^\frac{1}{3}\\
&+\frac{2rkmL^2}{72\kappa_{Tr}^\phi(\w)}\\
=&\left(-\frac{8r^3k^3m^3L^9+27\epsilon}{27L^3\kappa_{Tr}^\phi(\w)}+\left(-\frac{2^7}{27}\frac{r^6k^6m^6L^{3}}{\kappa_{Tr}^\phi(\w)^6}+\frac{r^6\epsilon^2}{L^6\kappa_{Tr}^\phi(\w)^2}-\frac{2^4r^6\epsilon k^3m^3L^{\frac32}}{27\kappa_{Tr}^\phi(\w)^4}\right)^\frac{1}{2}\right)^\frac{1}{3} \\
&+ \left(-\frac{8r^3k^3m^3L^9+27\epsilon}{27L^3\kappa_{Tr}^\phi(\w)}-\left(-\frac{2^7}{27}\frac{r^6k^6m^6L^{3}}{\kappa_{Tr}^\phi(\w)^6}+\frac{r^6\epsilon^2}{L^6\kappa_{Tr}^\phi(\w)^2}-\frac{2^4r^6\epsilon k^3m^3L^{\frac32}}{27\kappa_{Tr}^\phi(\w)^4}\right)^\frac{1}{2}\right)^\frac{1}{3} \\
&+\frac{2rkmL^2}{72\kappa_{Tr}^\phi(\w)}\\
    \end{split}
\end{equation*}
\end{proof}

\subsection{Derivation of Hessian and Third Derivative}
\label{app:hessian}
Let $\phi \in \R^{m}$ denote the embedding of the feature extractor and $W\in \R^{K \times m}$, where we denote the weights of the 
$k$-th neuron as $w_k$. The output layer is given by the softmax function $\pred_k = \operatorname{softmax}(W\phi) \in \R^K$.
More precisely the softmax is given by,

$$ \pred_k = \frac{\exp(w_k \phi)}{\sum_{j=1}^K \exp(w_j \phi)}$$

For simplicity, we omit the bias term. The one-hot encoded ground truth is given by $y$. 
The derivative of the loss $L$ function wrt. the weight vector $w_j$ can be computed as

$$ \frac{\partial L(y,\hat{y})}{\partial w_j} = - (y_j - \pred_j)\phi^T$$

\paragraph{Second derivative}

\begin{align*}
    \frac{\partial L(y,\hat{y})}{\partial w_l \; \partial w_j} &= \frac{\partial}{\partial w_l}- (y_j - \pred_j)\phi^T \\
                                                               &= \partl{l} - y_j\phi^T  +  \partl{l} \pred_j     \phi^T  \\
                                                               &= \partl{l} \pred_j     \phi^T  
\end{align*} 
The last equation follows from $y$ being independent of $w_l$. Next, we do a case analysis on $l=j$.
\begin{itemize}
    \item ( $l=j$): In (1), we use the quotient rule and in (2) definition of softmax. \begin{align*} 
                        \partl{l} \pred_j   \phi^T  &=  \partl{j} \frac{\exp(w_j \phi)}{\sum_{k=1}^K \exp(w_k \phi)} \phi^T \\
                                                  &= \frac{(\exp(w_j \phi) \sum_{k=1}^K \exp(w_k \phi)) \phi \phi^T - \exp(w_j \phi) \exp(w_j \phi) \phi \phi^T }{(\sum_{k=1}^K \exp(w_k \phi))^2} \; \text{(1)}  \\
                                                  &= \left (\frac{\exp(w_j \phi) \sum_{k=1}^K \exp(w_k \phi) }{(\sum_{k=1}^K \exp(w_k \phi))^2} -\frac{\exp(w_j \phi)^2 }{(\sum_{k=1}^K \exp(w_k \phi))^2}  \right)\phi \phi^T \\
                                                  &= (\pred_j - \pred_j^2)\phi \phi^T \in \R^{m\times m} \; \text{(2)}
                    \end{align*} 
     \item ( $l\neq j$): Again quotient rule, but the left side vanishes.
     \begin{align*}
          \partl{l} \pred_j   \phi^T  = - \pred_l \pred_j \phi \phi^T \in \R^{m\times m}
     \end{align*}
\end{itemize}
Then we have
$$\frac{\partial L(y,\hat{y})}{\partial w_l \; \partial w_j} = \pred_l (\mathbbm{1}_{[l=j]}- \pred_j) \phi \phi^T \in \R^{m\times m}$$

The hessian is then given by
$$ H(L;W)(y,\pred)=(\text{diag}(\pred) - \pred \pred^T  ) \otimes \phi \phi^T \;\in \R^{Km\times Km}  $$

\paragraph{Third derivative}
First, rewrite 
\begin{align*}
    \frac{\partial L(y,\hat{y})}{\partial w_l \; \partial w_j} = \pred_l (\mathbbm{1}_{[l=j]}- \pred_j) \phi \phi^T =  \pred_l \mathbbm{1}_{[l=j]} \phi \phi^T - \pred_l \pred_j \phi \phi^T
\end{align*}
Then we define a new operator $ \pro: \R^n \times \R^m \times \R^o \rightarrow \R^{n\times m \times o}, \pro(x,y,z)_{ijk} = x_iy_j z_k$. 
We now compute 
 $$\frac{\partial L(y,\hat{y})}{\partial w_o\,\partial w_l \, \partial w_j} = \partl{o}  \pred_l \mathbbm{1}_{[l=j]} \phi \phi^T - \pred_l \pred_j \phi \phi^T$$

Again we make a CA on $j=l$
\begin{itemize}
    \item ( $l=j$): \begin{align*}
        \partl{o} ( \pred_l \phi \phi^T - \pred_l ^2 \phi \phi^T )&= \partl{o}  \pred_l \phi \phi^T - \partl{o}  \pred_l ^2 \phi \phi^T \\
                                                                  &= \pred_o (\mathbbm{1}_{[o=l]}- \pred_l) \pro(\phi,\phi,\phi) - 2   (\pred_o (\mathbbm{1}_{[o=l]}- \pred_l) \pro(\phi,\phi,\phi
                                                                  )) \\
                                                                  &=  - \pred_o (\mathbbm{1}_{[o=l]}- \pred_l) \pro(\phi,\phi,\phi)
    \end{align*}
    \item ( $l\neq j$):
    \begin{align*}
         \partl{o} - \pred_l \pred_j \phi \phi^T &= - (\partl{o} \pred_l) \pred_j \phi \phi^T -  \pred_l (\partl{o} \pred_j)\phi \phi^T \\
                                                 &=  - \pred_j \pred_o (\mathbbm{1}_{[o=l]}- \pred_l) \cdot \pro(\phi,\phi,\phi) -  \pred_l \pred_o(\mathbbm{1}_{[o=i]}- \pred_j) \cdot \pro(\phi,\phi,\phi) \\
                                                 &=  - [\pred_j \pred_o (\mathbbm{1}_{[o=l]}- \pred_l)  +  \pred_l \pred_o(\mathbbm{1}_{[o=i]}- \pred_j) ] \cdot \pro(\phi,\phi,\phi) \in \R^{m\times m\times m}
    \end{align*}
    $\rightarrow - [\pred_j \pred_o (\mathbbm{1}_{[o=l]}- \pred_l)  +  \pred_l \pred_o(\mathbbm{1}_{[o=j]}- \pred_j) ]_{j,l,o=1..k} \otimes \pro(\phi,\phi,\phi) \in \R^{Km\times Km\times Km}$
\end{itemize}

\section{Additional results}
\label{app:add-results}
\subsection{Results for other architectures and datasets}
We provide the remaining results in Figure \ref{app:resnet-losses} to Figure \ref{fig:dense-losses}. We see the same trend as 
observed in the main paper. The only difference is that for models trained CIFAR-100 the basin width correlates not 
as strong with the sharpness at 0 compared to models trained CIFAR-10.
\input{latex_figs/new_figs/vgg_losses.tex}
\input{latex_figs/new_figs/resnet_losses.tex}
\input{latex_figs/new_figs/wrn_losses.tex}
\input{latex_figs/new_figs/dense_losses.tex}

\subsection{Evaluation on adversarially trained ResNet}
We also run the same evaluation as in Section \ref{sec:eval} with an adversarially trained ResNet. For training, we use standard adversarially training with PGD-$\ell_\infty$ (10 steps, $\epsilon = 8/255$, step size $\alpha = 2/255$). Given that this will result in a more robust model, we use a stronger attack as in the original evaluation, namely PGD-$\ell_2$ (50 steps, $\epsilon = 0.5$, step size $\alpha = 0.01$). This attack explores a radius that $20$ times bigger than in the original evaluation. We present the results in Figure \ref{app:adv-resnet-losses}. In general, we observe exactly the same trend as for all other models and datasets. The main difference is that the basin width is way bigger, as stated above $20\times$.
This means the classification will be constant in a larger radius in the input space, yet the curvature as measured by $tr(H)$  is similar e.g $0.6$ for clean and $1.0$ for adversarial.
The main difference stems from the fact that adversarially trained networks are less confident.
\begin{figure}[h!]
    \begin{subfigure}[b]{0.25\linewidth}
        \begin{tikzpicture}
            \begin{axis}[
                xlabel={Loss increase},
                ylabel={Frequency},
                pretty ybar,
                cycle list name = prcl-ybar,
                height=3cm,
                width=1.1\linewidth,
                xmin=0, xmax=4,
                xtick={0,1,2,3,4},
                legend style={yshift=-0.1cm,,xshift=-0.8cm,font=\scriptsize},
                legend style={font=\tiny},
                legend entries = {$s=0.25$,$s=1$,$s=10$},
                legend columns = 1,
                ylabel style={at={(axis description cs:-0.3,.5)},anchor=south},
                yticklabel style={xshift=2pt},
            ]

            \addplot +[
                hist={
                    bins=30,
                    },opacity = 0.25, red,
                    x filter/.code={
                    \pgfmathparse{(\pgfmathresult < 4) ? \pgfmathresult : nan}%
                    \let\pgfmathresult=\pgfmathresult},
            ] table [y index=1,col sep=comma] {new_results/losses/resnet-eps-8/scale_0.25_resnet-eps-8_analysis_results.csv};

            \addplot +[
            hist={bins=30},
            opacity=0.25,
            %
            x filter/.code={
                \pgfmathparse{(\pgfmathresult < 4) ? \pgfmathresult : nan}%
                \let\pgfmathresult=\pgfmathresult
            },
            ]%
            table[y index=1, col sep=comma] {new_results/losses/resnet-eps-8/scale_1_resnet-eps-8_analysis_results.csv};

            \addplot +[
            hist={bins=30},
            opacity=0.25,
            %
            x filter/.code={
                \pgfmathparse{(\pgfmathresult < 4) ? \pgfmathresult : nan}%
                \let\pgfmathresult=\pgfmathresult
            },
            ]%
            table[y index=1, col sep=comma] {new_results/losses/resnet-eps-8/scale_10_resnet-eps-8_analysis_results.csv};

        \end{axis}
        \end{tikzpicture}
        \subcaption{}
    \end{subfigure}%
    \begin{subfigure}[b]{0.25\linewidth}
            \begin{tikzpicture}[baseline=(current axis.outer south)]
                    \usetikzlibrary{calc}
                    \begin{axis}[ 
                        smalljonas line,
                        cycle list name = prcl-line,
                        clip=false, 
                        legend style={
                            at={(0.5,1.02)}, anchor=south,    
                            font=\scriptsize
                          },
                        legend image post style={xscale=0.55},
                        width=\linewidth,
                        height=3cm,
                        ylabel={Normalized Loss},
                        xlabel={Iterations},
                        pretty labelshift,
                        line width=0.75,
                        xtick={0,10,20,30,40,50},
                        ymax=1,
                        ymin=0.0,xmin=0.0,xmax=50,
                        legend columns = 3,
                        tick label style={/pgf/number format/fixed},
                        legend entries = {Adv, Clean}
                    ]
    
                    \pgfplotsinvokeforeach{0.5,1,2.5,5,10,50}{%
                            \addplot+[
                                mark=*,
                            ] table[
                                y = normalized_losses,   
                                col sep = comma,
                                x expr = \coordindex + 1 
                            ] {new_results/losses/other_basins/resnet-eps-8_#1.csv};
                            \addlegendentry{#1}          
                        }%

                    \end{axis}
                \end{tikzpicture}
                \subcaption{}
                \label{sub:basin}
    \end{subfigure}%
    \begin{subfigure}[b]{0.25\linewidth}
        \begin{tikzpicture}
                \usetikzlibrary{calc}
                \begin{axis}[ 
                    smalljonas line,
                    cycle list name = prcl-line,
                    legend style={yshift=-0.4cm,,xshift=-0.8cm,font=\scriptsize},
                    width=\linewidth,
                    height=3cm,
                    ylabel={Loss},
                    xlabel={Iterations},
                    pretty labelshift,
                    line width=0.75,
                    restrict y to domain*=0:52,
                    unbounded coords = jump,
                    clip mode=individual,
                    ymax=50,
                    ymin=0.0,
                    xmax=25,
                    legend columns = 1,
                    ytick={0,10,20,30,40,50},xtick={0,0,5,10,15,20,25},xmin=0,
                    tick label style={/pgf/number format/fixed},
                ]
                \pgfplotsinvokeforeach{loss_0,loss_1,loss_2,loss_3,loss_4}{
                    \addplot+[
                        mark=*,
                    ] table[
                        y = #1,   
                        col sep = comma,
                        x expr = \coordindex + 1 
                    ] {new_results/losses/resnet-eps-8/basins_resnet-eps-8_analysis_results.csv};         
                }
                \addplot+[
    domain=1:50,  
    samples=2,     
    color=white,
    line width=4pt, opacity=1 
] {52};
                \end{axis}
            \end{tikzpicture}
                    \subcaption{}
    \end{subfigure}%
    \begin{subfigure}[b]{0.25\linewidth}
        \begin{tikzpicture}
                \usetikzlibrary{calc}
                \begin{axis}[ 
                    pretty scatter,
                    legend style={yshift=-0.4cm,,xshift=-0.8cm,font=\scriptsize},
                    width=\linewidth,
                    height=3cm,
                    ylabel={Basin width},
                    xlabel={Relative Sharpness@0},
                    pretty labelshift,
                    line width=0.75,
                    ymin=0.0, ymax=50,
                    xmax=750,
                    legend columns = 1,
                    ytick={0,10,20,30,40,50},
                    xmin=0,
                    tick label style={/pgf/number format/fixed},
                ]
                    \addplot[only marks, mark size=1.2pt]
                    table[x={sharpness_at_zeros}, y={take_off}, col sep=comma,x filter/.code=\ifnum\coordindex>1200\def\pgfmathresult{}\fi]{new_results/losses/resnet-eps-8/take_off_resnet-eps-8_analysis_results.csv};
                \end{axis}
            \end{tikzpicture}
                    \subcaption{}
\end{subfigure}%
    
    \caption{\textit{Loss geometry for} \resnet .\textbf{(a)} We report the distribution of the loss increase for varying scaling values.
    \textbf{(b)} We show for one example how the basin formes as we increase the scaling. We use the normalized loss to plot all in one axis.
    \textbf{(c)} We show examples of samples exhibiting different basin widths.
    \textbf{(d)} We plot the sharpness measured at the test data ($s=1$) and the basin width. }
    \label{app:adv-resnet-losses}
    \end{figure}

\subsection{Flatness can induce gradient masking}
To test whether the geometry backpropagates to the input, we need to compute the Hessian with respect to the input,
which is prohibitively expensive. Instead, we take an actionable route: if the loss surface is truly flat,
then first-order attacks must fail, as without curvature the optimization cannot succeed i.e. they become unattackable.
We attack ResNet-18 models with standard PGD-$\ell_\infty$ (10 steps, $\epsilon = 8/255$, step size $\alpha = 2/255$) while
varying the penultimate layer scaling factor $s$ from 1 to 100. We measure the robust test accuracy and average sharpness at the clean examples
and show the results in Figure \ref{fig:unattackable}.

As we can see, increasing $s$ dramatically improves robust test accuracy. The standard ResNet with $s=1$ achieves only 0\% robust accuracy, while its scaled counterpart with $s=100$ reaches 93\%, which recovers the clean accuracy. Crucially, the relative sharpness measured at the penultimate layer strongly correlates with robust test accuracy across all scaling values. This direct correlation demonstrates that penultimate layer geometry indeed propagates to the input space. 
Importantly, this does not mean that scaled networks are adversarially robust to according to Defintion \ref{def:datasetRobust}Instead, the flat geometry prevents first-order attacks from finding adversarial examples.
More specifically, adversarial examples found at $s=1$ transfer with 100\% success to other scales, indicating that the same vulnerabilities remain but are harder to reach.
\emph{This highlights that analyzing adversarial robustness through sharpness is more nuanced than it appears and requires careful evaluation.}
\begin{figure}[h]
\begin{subfigure}[h]{0.25\linewidth}
        \begin{tikzpicture}
                \usetikzlibrary{calc}
                \begin{axis}[ 
                    smalljonas line,
                    cycle list name = prcl-line,
                    legend style={yshift=1.0cm,font=\scriptsize,xshift=4cm},
                    width=\linewidth,
                    height=3cm,
                    ylabel={Relative sharpness},
                    xlabel={Scaling $s$},
                    pretty labelshift,
                    line width=0.75,
                    ymin=0.0,xmin=0.0,
                    legend columns = 9,
                    tick label style={/pgf/number format/fixed},
                ]
                    \addplot
                    table[x={scales}, y={sharpness_at_zero}, col sep=comma]{new_results/unattackable/resnet-eps-0_analysis_results.csv};
                    \addplot[black, dashed, domain=0:100] {0};
                \end{axis}
            \end{tikzpicture}
                    \subcaption{}
\end{subfigure}%
\begin{subfigure}[h]{0.25\linewidth}
        \begin{tikzpicture}
                \usetikzlibrary{calc}
                \begin{axis}[ 
                    smalljonas line,
                    cycle list name = prcl-line,
                    legend style={yshift=-0.4cm,,xshift=0cm,font=\scriptsize},
                    width=\linewidth,
                    height=3cm,
                    ylabel={Robust Test Acc.},
                    xlabel={Scaling $s$},
                    pretty labelshift,
                    line width=0.75,
                    ymax=100,
                    ymin=0.0,xmin=0.0,
                    legend columns = 1,
                    tick label style={/pgf/number format/fixed},
                    legend entries = {Adv, Clean}
                ]

                    \addplot
                    table[x={scales}, y={adv_accs_1}, col sep=comma]{new_results/unattackable/resnet-eps-0_analysis_results.csv};
                    \addplot[black, dashed, domain=0:100] {94.5};
                \end{axis}
            \end{tikzpicture}
                    \subcaption{}
\end{subfigure}%
\begin{subfigure}[h]{0.25\linewidth}
        \begin{tikzpicture}
                \usetikzlibrary{calc}
                \begin{axis}[ 
                    smalljonas line,
                    cycle list name = prcl-line,
                    legend style={yshift=-0.4cm,,xshift=-0.8cm,font=\scriptsize},
                    width=\linewidth,
                    height=3cm,
                    ylabel={Robust Test Acc.},
                    xlabel={Relative Sharpness},
                    pretty labelshift,
                    line width=0.75,
                    ymax=100,
                    ymin=0.0,
                    legend columns = 1,
                    tick label style={/pgf/number format/fixed}, x dir=reverse,
                    legend entries = {Adv, Clean}
                ]
                    \addplot
                    table[x={sharpness_at_zero}, y={adv_accs_1}, col sep=comma]{new_results/unattackable/resnet-eps-0_analysis_results.csv};
                    \addplot[black, dashed, domain=0:12] {94.5};
                \end{axis}
            \end{tikzpicture}
                    \subcaption{}
\end{subfigure}%
\begin{subfigure}[h]{0.25\linewidth}
    \begin{tikzpicture}
            \usetikzlibrary{calc}
            \begin{axis}[ 
                smalljonas line,
                cycle list name = prcl-line,
                legend style={yshift=-0.4cm,,xshift=-0.8cm,font=\scriptsize},
                width=\linewidth,
                height=3cm,
                ylabel={Robust Test Acc.},
                xlabel={Relative Sharpness},
                pretty labelshift,
                line width=0.75,
                ymax=100,
                ymin=0.0,
                legend columns = 1,
                tick label style={/pgf/number format/fixed}, x dir=reverse,
                legend entries = {Adv, Clean}
            ]
                \addplot [dollarbill, domain=0:12] {0};
                \addplot[black, dashed, domain=0:12] {94.5};
            \end{axis}
        \end{tikzpicture}
                \subcaption{}
                \label{fig:transfer}
\end{subfigure}%

\caption{\textit{Unattackable networks.} \textbf{(a)} we report the relative sharpness of the scaled networks at the test data.
\textbf{(b)} Robust Test Accuracy with varying scaling $s$. \textbf{(c)} relative sharpness at the clean data vs. Robust Test Accuracy.
\textbf{(d)} transferability of the attacks created on $s=1$. }
\label{fig:unattackable}
\end{figure}
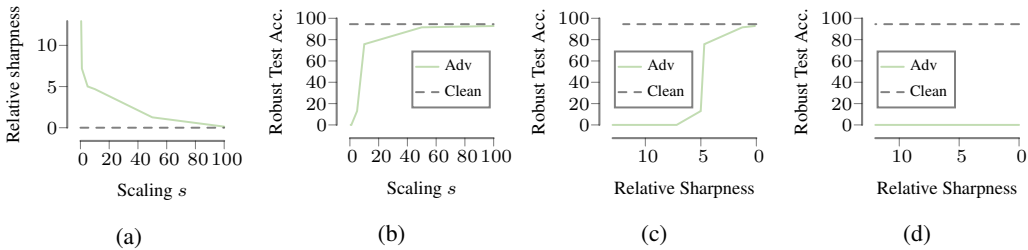
\begin{figure}[h!]
\begin{subfigure}{0.2\linewidth}
        \begin{tikzpicture}
                \usetikzlibrary{calc}
                \begin{axis}[ 
                    smalljonas line,
                    cycle list name = more-prcl,
                    legend style={yshift=1cm,font=\scriptsize,xshift=9cm},
                    width=1.08\linewidth,
                    height=2.9cm,
                    ylabel={Relative flatness},
                    xlabel={Attack iteration},
                    pretty labelshift,
                    line width=0.75,
                    ymin=0,
                    ymax=50,
                    xmin=0,
                    xmax=10,
                    legend columns = 9,
                    tick label style={/pgf/number format/fixed},
                ]
                    \addplot table[x={0},y={layer_103},col sep=comma]{expres/sharpness_per_layer.csv};%

                \end{axis}
            \end{tikzpicture}
            \subcaption{Layer $l$}
\end{subfigure}%
\begin{subfigure}{0.2\linewidth}
        \begin{tikzpicture}
                \usetikzlibrary{calc}
                \begin{axis}[ 
                    smalljonas line,
                    cycle list name = more-prcl,
                    legend style={yshift=1cm,font=\scriptsize,xshift=9cm},
                    width=1.08\linewidth,
                    height=2.9cm,
                    ylabel={Relative flatness},
                    xlabel={Attack iteration},
                    pretty labelshift,
                    line width=0.75,
                    ymin=0,
                    ymax=240,
                    xmin=0,
                    xmax=10,
                    legend columns = 9,
                    tick label style={/pgf/number format/fixed},
                ]
                    \addplot table[x={0},y={layer_100},col sep=comma]{expres/sharpness_per_layer.csv};%

                \end{axis}
            \end{tikzpicture}
            \subcaption{Layer $l-1$}
\end{subfigure}%
\begin{subfigure}{0.2\linewidth}
        \begin{tikzpicture}
                \usetikzlibrary{calc}
                \begin{axis}[ 
                    smalljonas line,
                    cycle list name = more-prcl,
                    legend style={yshift=1cm,font=\scriptsize,xshift=9cm},
                    width=1.08\linewidth,
                    height=2.9cm,
                    ylabel={Relative flatness},
                    xlabel={Attack iteration},
                    pretty labelshift,
                    line width=0.75,
                    ymin=0,
                    ymax=800,
                    xmin=0,
                    xmax=10,
                    legend columns = 9,
                    tick label style={/pgf/number format/fixed},
                ]
                    \addplot table[x={0},y={layer_97},col sep=comma]{expres/sharpness_per_layer.csv};%

                \end{axis}
            \end{tikzpicture}
                        \subcaption{Layer $l-2$}
                        \label{fig:adv-train-c-layers}
\end{subfigure}%
\begin{subfigure}{0.2\linewidth}
        \begin{tikzpicture}
                \usetikzlibrary{calc}
                \begin{axis}[ 
                    smalljonas line,
                    cycle list name = more-prcl,
                    legend style={yshift=1cm,font=\scriptsize,xshift=9cm},
                    width=1.08\linewidth,
                    height=2.9cm,
                    ylabel={Relative flatness},
                    xlabel={Attack iteration},
                    pretty labelshift,
                    line width=0.75,
                    ymin=0,
                    ymax=700,
                    xmin=0,
                    xmax=10,
                    legend columns = 9,
                    tick label style={/pgf/number format/fixed},
                ]
                    \addplot table[x={0},y={layer_94},col sep=comma]{expres/sharpness_per_layer.csv};%

                \end{axis}
            \end{tikzpicture}
                        \subcaption{Layer $l-3$}
\end{subfigure}%
\begin{subfigure}{0.2\linewidth}
        \begin{tikzpicture}
                \usetikzlibrary{calc}
                \begin{axis}[ 
                    smalljonas line,
                    cycle list name = more-prcl,
                    legend style={yshift=1cm,font=\scriptsize,xshift=9cm},
                    width=1.08\linewidth,
                    height=2.9cm,
                    ylabel={Relative flatness},
                    xlabel={Attack iteration},
                    pretty labelshift,
                    line width=0.75,
                    ymin=0,
                    ymax=1500,
                    xmin=0,
                    xmax=10,
                    legend columns = 9,
                    yticklabels = {0,0,500, 1k, 1.5k},
                    tick label style={/pgf/number format/fixed},
                ]
                    \addplot table[x={0},y={layer_91},col sep=comma]{expres/sharpness_per_layer.csv};%

                \end{axis}
            \end{tikzpicture}
                        \subcaption{Layer $l-4$}
\end{subfigure}
\caption{We show the relative sharpness measure computed in the penultimate layer $l$ and in shallower layers $l-1$ to $l-4$ for
\wrn. Due to memory and runtime constraints, we
approximate the measure using Hutchinson trace estimation used in \cite{petzka2021relative} on 500 images. We observe
the same phenomena as in the penultimate layer, which justifies that we focus only on the penultimate layer for our theoretical
and experimental analysis.}
\label{fig:diff-layer}
\end{figure}
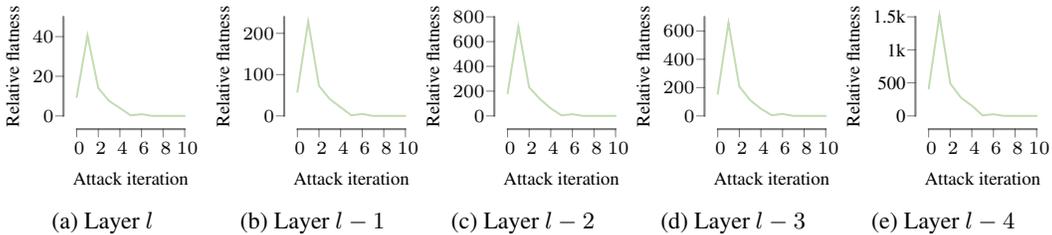
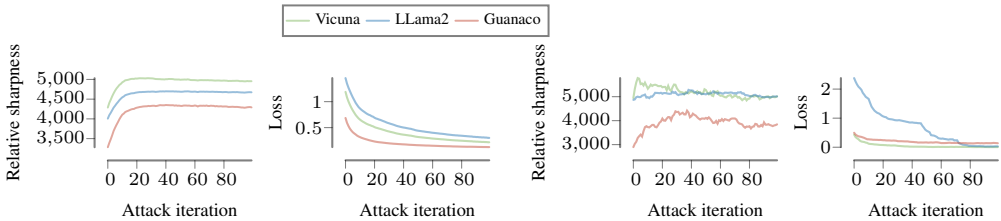
\begin{figure}[h!]
\begin{subfigure}{0.25\linewidth}
        \begin{tikzpicture}
                \usetikzlibrary{calc}
                \begin{axis}[ 
                    smalljonas line,
                    cycle list name = prcl-line,
                    legend style={yshift=1cm,font=\scriptsize,xshift=4cm},
                    width=\linewidth,
                    height=2.5cm,
                    ylabel={Relative sharpness},
                    xlabel={Attack iteration},
                    pretty labelshift,
                    line width=0.75,
                    legend columns = 9,
                    tick label style={/pgf/number format/fixed},
                    legend entries = {Vicuna, LLama2, Guanaco},
                ]
                \foreach \x in {vicuna, llama2, guanaco}{
                    \addplot table[x={index},y={\x-fam},col sep=comma]{expres/llms/\x-attack.csv};%
                }    
                \end{axis}
            \end{tikzpicture}
\end{subfigure}%
\begin{subfigure}{0.25\linewidth}
        \begin{tikzpicture}
                \usetikzlibrary{calc}
                \begin{axis}[ 
                    smalljonas line,
                    cycle list name = prcl-line,
                    legend style={yshift=0.7cm,font=\scriptsize,xshift=4cm},
                    width=\linewidth,
                    height=2.5cm,
                    ylabel={Loss},
                    xlabel={Attack iteration},
                    pretty labelshift,
                    legend columns = 5,
                    tick label style={/pgf/number format/fixed}
                ]
                \foreach \x in {vicuna, llama2, guanaco}{
                    \addplot table[x={index},y={\x-loss},col sep=comma]{expres/llms/\x-attack.csv};%
                }       
                \end{axis}
            \end{tikzpicture}
                \label{fig:llm-adv-b}
\end{subfigure}%
\begin{subfigure}{0.25\linewidth}
        \begin{tikzpicture}
                \usetikzlibrary{calc}
                \begin{axis}[ 
                    smalljonas line,
                    cycle list name = prcl-line,
                    legend style={yshift=0.7cm,font=\scriptsize,xshift=3.4cm},
                    width=\linewidth,
                    height=2.5cm,
                    ylabel={Relative sharpness},
                    xlabel={Attack iteration},
                    pretty labelshift,
                    legend columns = 5,
                    tick label style={/pgf/number format/fixed}
                ]
                \foreach \x in {vicuna, llama2, guanaco}{
                    \addplot table[x={index},y={fam_pos},col sep=comma]{expres/posneg/\x_posneg.csv};%
                }       
                \end{axis}
            \end{tikzpicture}
            \label{fig:llm-adv-c}
\end{subfigure}%
\begin{subfigure}{0.25\linewidth}
        \begin{tikzpicture}
                \usetikzlibrary{calc}
                \begin{axis}[ 
                    smalljonas line,
                    cycle list name = prcl-line,
                    legend style={yshift=0.7cm,font=\scriptsize,xshift=3.4cm},
                    width=\linewidth,
                    height=2.5cm,
                    ylabel={Loss},
                    xlabel={Attack iteration},
                    pretty labelshift,
                    legend columns = 5,
                    tick label style={/pgf/number format/fixed}
                ]
                \foreach \x in {vicuna, llama2, guanaco}{
                    \addplot table[x={index},y={loss_pos},col sep=comma]{expres/posneg/\x_posneg.csv};%
                }       
                \end{axis}
            \end{tikzpicture}
            \label{fig:llm-adv-d}
\end{subfigure}
\caption{\textbf{Top}. We plot the relative sharpness and loss of the adversarial prompt for Viucna, LLama2 and Guanaco when attacked by
the method of \cite{zou2023universal}. \textbf{Bottom}.We give per model example trajectories that  first get sharper and then flatter again.}
\label{fig:llm-adv}
\end{figure}

\subsection{Detecting Adversarial examples}
\label{app:detection}
It is possible to detect adversarial examples using a simple threshold on the relative sharpness measure. We did not include a practical study of this since it would go beyond the scope of this paper: Developing a sound method requires more than fine-tuning the threshold. Moreover, this requires comparing the approach to a wide range of state-of-the-art detection methods, which would be a paper of its own. Therefore, we leave this interesting practical aspect for future work. Nonetheless, we provide preliminary results. We  trained a decision stump on the sharpness of clean and adversarial samples on CIFAR-10 for \wrn using 5-fold cross-validation, which yields the following accuracies: $[0.92, 0.92, 0.93, 0.92, 0.92] $, i.e., adversarial examples can be detected with an average accuracy of $0.92$ with little to no difference between the folds.

\section{Related Work}\label{app:rlw}
\paragraph{Defenses and Guarantees}
The most widely used method for improving robustness to adversarial examples is adversarial training, which integrates adversarial perturbations into the training process~\citep{szegedy2014intriguing, shafahi2019adversarial, kumari2019harnessing, perolat_playing_2018, shafahi2020universal, cai2018curriculum, tramer2018ensemble, wu2020adversarial, carmon2019unlabeled}. 
While this approach enhances adversarial robustness, it frequently leads to a reduction in clean accuracy\newline\citep{tsipras2018robustness, rade2021helper, nakkiran2019adversarial, zhang2019theoretically}. 
This trade-off remains controversial, prompting the question of whether it is fundamentally impossible to simultaneously achieve high performance and robustness.

One hypothesis is that adversarial training is inherently more difficult than standard training, and that networks simply underperform when faced with this more challenging task~\citep{schmidt2018adversarially}.
Conversely, \citet{raghunathan2020understanding} argue that deep neural networks are capable of fitting even random noise~\citep{zhang2017understanding}, suggesting that insufficient regularization, rather than task complexity, may be the root cause of this performance drop.
Indeed, numerous works have explored strategies to improve generalization and thereby enhance the effectiveness of adversarial training.
These include incorporating additional data~\citep{carmon2019unlabeled, alayrac2019labels, hendrycks2019using, gowal2021improving}, using early stopping~\citep{rice2020overfitting}, and promoting flatness in the loss landscape~\citep{wu2020adversarial, stutz_relating_2021, moosavi2019robustness, xu2020adversarial}.

To better understand this controversy and nature of adversarial vulnerability, a line of research has proposed the \emph{in- and off-manifold hypothesis}~\citep{stutz2019disentangling, shamir2021dimpled, zhang2022manifold, haldar2024effect, melamed2023adversarial}.
According to this perspective, adversarial examples can be either on the data manifold or off it.
Robust generalization is achievable primarily for in-manifold adversarial examples, while off-manifold ones tend to degrade clean accuracy during adversarial training.
\citet{song2018pixeldefend} support this view empirically, proposing a defense mechanism that learns the data manifold and projects adversarial examples onto it, thereby restoring correct classification.

This perspective requires additional assumptions about the data distribution.
In practice, we often have access to only a finite and relatively small dataset---especially when compared to the high dimensionality of input spaces---which limits our ability to precisely model or leverage the underlying manifold structure.

From the other side, a natural approach to controlling model behavior on unseen samples is through the Lipschitz constant: if an input is close to a training sample, the model’s prediction should not vary significantly. 
However, applying Lipschitz constraints directly to modern deep networks proves challenging. Global Lipschitz constants~\citep{tsuzuku2018lipschitz} tend to be too coarse to be meaningful, while local Lipschitz constants~\citep{hein2017formal} are computationally intractable to estimate accurately.
Moreover, \citet{liang2021large} show that a large global Lipschitz constant does not necessarily imply large local constants, meaning that a model may still be locally robust despite a high global value. 
Conversely, enforcing a small global Lipschitz constant can help control local behavior but often leads to degraded clean accuracy. 
\citet{yang2020closer} suggest that improving local Lipschitz constants alongside generalization could yield both robust and accurate models, though achieving this balance is not straightforward in practice.

Efforts to explicitly control Lipschitz properties during training have inspired the development of certification-based robustness techniques.
In these methods, models are trained to produce consistent predictions under random input perturbations, which yields certified guarantees of robustness on training samples~\citep{cohen2019certified, salman2019provably}.
Interestingly, \citet{kanai2023relationship} analyze robustness in both the input and parameter spaces, concluding that enforcing smoothness in the input space often results in a non-flat loss surface with respect to parameters, which in turn harms generalization.
This suggests that input-space robustness alone may not suffice to preserve clean accuracy, while parameter-space flatness alone may be insufficient for adversarial robustness.
This tension ties into the broader understanding of generalization: robustness in parameter space—measured via loss surface flatness—is known to correlate with improved generalization~\citep{hochreiter1994simplifying, liang2019fisher, tsuzuku2020normalized, foret2020sharpness, petzka2021relative}. 
Accordingly, enforcing smoothness in both the input and parameter spaces can lead to models that are both robust and generalize well, as demonstrated empirically by \citet{wu2020adversarial}, who also highlight the connection between weight flatness and the robust generalization gap.
Taking another perspective on measuring smoothness, \citet{simon2019first} show that adversarial examples tend to correspond to regions with higher input gradients, further linking gradient norm regularization to robustness. 
However, this line of research typically does not address whether such regularization helps against in-manifold or off-manifold adversarial examples.

In our analysis, we connect flatness in parameter space, network Lipschitz properties, and adversarial robustness.
We argue that while these defenses offer meaningful local guarantees, achieving global robustness necessitates additional assumptions about the structure and properties of the data manifold.
\citet{xuan2025exploring} came to similar to our conclusions with respect to the effect of network confidence on adversarial robustness, but through adjusting temperature.
They show empirically in preliminary experiments that high temperature in cross-entropy loss promotes better adversarial robustness.
\citet{xu2024understanding} derives a measure bounding robustness generalization gap using PAC-Bayes bounds.
This measure surprisingly repeats the relative flatness~\citep{petzka2021relative} that we use in our derivation, confirming that such theoretical connection can be proven from different perspectives.
\section{Unnormalized Plots}
\label{app:plots}
\begin{figure}[h!]
\centering
\begin{subfigure}[h]{0.4\linewidth}
        \begin{tikzpicture}
                \usetikzlibrary{calc}
                \begin{axis}[ 
                    smalljonas line,
                    cycle list name = prcl-line,
                    legend style={yshift=1.0cm,font=\scriptsize,xshift=4cm},
                    width=\linewidth,
                    height=3cm,
                    ylabel={Relative sharpness},
                    xlabel={Attack iteration},
                    pretty labelshift,
                    line width=0.75,
                    legend columns = 9,
                    tick label style={/pgf/number format/fixed},
                    legend entries = {\wrn,\resnet,\vgg,\dense}
                ]
                \foreach \x in {wrn, resnet,vgg11_bn,dense121}{
                    \addplot+[forget plot, eda errorbarcolored, y dir=plus, y explicit]
                    table[x={0}, y={trace-\x-cifar10}, y error expr=\thisrow{trace-\x-cifar10}-\thisrow{stdtrace-\x-cifar10}, col sep=comma]{expres/debug-rfm-for-clean.csv};
                    \addplot+[eda errorbarcolored, y dir=minus, y explicit]
                    table[x={0}, y={trace-\x-cifar10}, y error expr=\thisrow{trace-\x-cifar10}-\thisrow{stdtrace-\x-cifar10}, col sep=comma]{expres/debug-rfm-for-clean.csv};
                }    
                \end{axis}
            \end{tikzpicture}
                    \subcaption{Sharpness CIFAR-10}
\end{subfigure}%
\begin{subfigure}[h]{0.4\linewidth}
        \begin{tikzpicture}
                \usetikzlibrary{calc}
                \begin{axis}[ 
                    smalljonas line,
                    cycle list name = prcl-line,
                    legend style={yshift=1.0cm,font=\scriptsize,xshift=6cm},
                    width=\linewidth,
                    height=3cm,
                    ylabel={Loss},
                    xlabel={Attack iteration},
                    pretty labelshift,
                    line width=0.75,
                    legend columns = 9,
                    tick label style={/pgf/number format/fixed},
                ]
                \foreach \x in {wrn, resnet,vgg11_bn,dense121}{
                    \addplot+[forget plot, eda errorbarcolored, y dir=plus, y explicit]
                    table[x={0}, y={loss-\x-cifar10}, y error expr=\thisrow{loss-\x-cifar10}-\thisrow{stdloss-\x-cifar10}, col sep=comma]{expres/debug-rfm-for-clean.csv};
                    \addplot+[eda errorbarcolored, y dir=minus, y explicit]
                    table[x={0}, y={loss-\x-cifar10}, y error expr=\thisrow{loss-\x-cifar10}-\thisrow{stdloss-\x-cifar10}, col sep=comma]{expres/debug-rfm-for-clean.csv};%
                }    
                \end{axis}
            \end{tikzpicture}
                    \subcaption{Loss CIFAR-10}
\end{subfigure}
\begin{subfigure}[h]{0.4\linewidth}
        \begin{tikzpicture}
                \usetikzlibrary{calc}
                \begin{axis}[ 
                    smalljonas line,
                    cycle list name = prcl-line,
                    legend style={yshift=1.0cm,font=\scriptsize,xshift=6cm},
                    width=\linewidth,
                    height=3cm,
                    ylabel={Relative sharpness},
                    xlabel={Attack iteration},
                    pretty labelshift,
                    line width=0.75,
                    legend columns = 9,
                    tick label style={/pgf/number format/fixed},
                ]
                \foreach \x in {wrn, resnet,vgg11_bn,dense121}{
                    \addplot+[forget plot, eda errorbarcolored, y dir=plus, y explicit]
                    table[x={0}, y={trace-\x-cifar100}, y error expr=\thisrow{trace-\x-cifar100}-\thisrow{stdtrace-\x-cifar100}, col sep=comma]{expres/debug-rfm-for-clean.csv};
                    \addplot+[eda errorbarcolored, y dir=minus, y explicit]
                    table[x={0}, y={trace-\x-cifar100}, y error expr=\thisrow{trace-\x-cifar100}-\thisrow{stdtrace-\x-cifar100}, col sep=comma]{expres/debug-rfm-for-clean.csv};%
                }    
                \end{axis}
            \end{tikzpicture}
            \subcaption{Sharpness CIFAR-100}
\end{subfigure}%
\begin{subfigure}[h]{0.4\linewidth}
        \begin{tikzpicture}
                \usetikzlibrary{calc}
                \begin{axis}[ 
                    smalljonas line,
                    cycle list name = prcl-line,
                    legend style={yshift=1.0cm,font=\scriptsize,xshift=6cm},
                    width=\linewidth,
                    height=3cm,
                    ylabel={Loss},
                    xlabel={Attack iteration},
                    pretty labelshift,
                    line width=0.75,
                    legend columns = 9,
                    tick label style={/pgf/number format/fixed},
                ]
                \foreach \x in {wrn, resnet,vgg11_bn,dense121}{
                    \addplot+[forget plot, eda errorbarcolored, y dir=plus, y explicit]
                    table[x={0}, y={loss-\x-cifar100}, y error expr=\thisrow{loss-\x-cifar100}-\thisrow{stdloss-\x-cifar100}, col sep=comma]{expres/debug-rfm-for-clean.csv};
                    \addplot+[eda errorbarcolored, y dir=minus, y explicit]
                    table[x={0}, y={loss-\x-cifar100}, y error expr=\thisrow{loss-\x-cifar100}-\thisrow{stdloss-\x-cifar100}, col sep=comma]{expres/debug-rfm-for-clean.csv};%
                }    
                \end{axis}
            \end{tikzpicture}
            \subcaption{Loss CIFAR-100}
\end{subfigure}
\caption{We report the relative sharpness on the attack trajectory of attack for \wrn, \resnet, \vgg and \dense
on the test set of CIFAR-10 \& CIFAR-100. We observe that adversarial examples first reach a sharp region, but with strength of the attack increasing they are in very flat region. We also display the standard deviation of the values on individual inputs.}
\label{fig:first}
\end{figure}
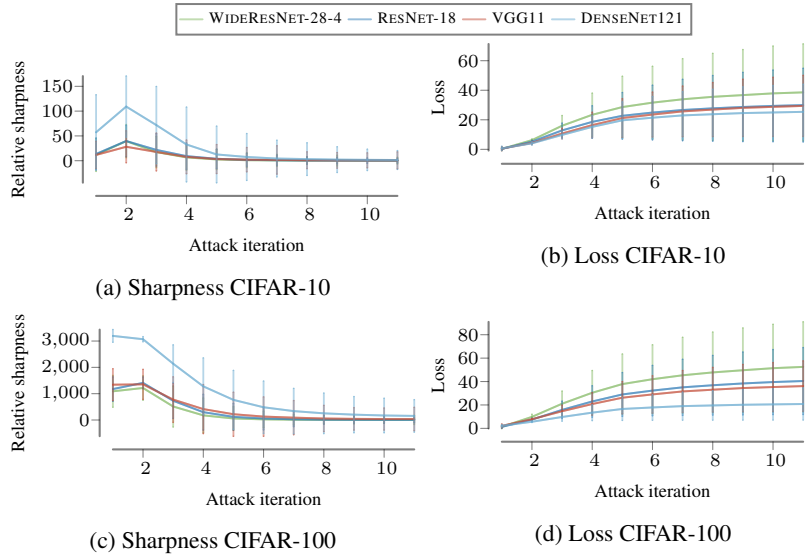

\section{Code for experiments}
We use code from several resources, which we disclose here. First, the basis for training and attacking the CNNs stems from \citep{sehwag2020hydra}.
We modified the code according to our needs. The code for DenseNet121 stems from the official PyTorch library. To attack and 
evaluate the LLMs, we use the official implementation of the attack \citep{zou2023universal}. CIFAR-10 and CIFAR-100 were also 
downloaded from PyTorch.

\section{LLM Use}
In this work, we used GPT-4o for both writing and coding support. On the writing side, it assisted with editing and condensing text to improve clarity. For coding, GPT-5 was used for debugging, providing autocomplete suggestions in VS Code, and generating code for LaTeX figures. Additionally, gpt-o3 was used to verify steps in the proofs and try to falsify certain steps.

\end{document}